\documentclass[runningheads]{llncs}

\usepackage{eccv}

\usepackage{eccvabbrv}

\usepackage{graphicx}
\usepackage{booktabs}

\usepackage{makecell}
\usepackage{todonotes}
\usepackage{tabularx}
\usepackage[table]{xcolor}
\usepackage{float}

\usepackage{multirow}
\usepackage{graphicx}

\usepackage[accsupp]{axessibility}  

\usepackage{hyperref}

\usepackage{orcidlink}

\input{Misc/header}
\begin{document}

\title{HSDF-Lane: Height-Aligned Signed Distance Field with Semantic Lane Prior for 3D Lane Detection} 


\titlerunning{HSDF-Lane}

\author{Jiyong Boo\orcidlink{0009-0000-6129-635X} \and
Byeongin Joung\orcidlink{0009-0004-0321-0430} \and
Hyemin Yang \orcidlink{0009-0002-2112-7804} \and
Kuk-Jin Yoon\thanks{Corresponding author} \orcidlink{0000-0002-1634-2756}}

\authorrunning{Boo et al.}

\institute{KAIST, Republic of Korea \\
\email{\{boojiyong, byeonginjoung, hyemin0806, kjyoon\}@kaist.ac.kr}}

\maketitle

\begin{abstract}

Monocular 3D lane detection plays a critical role in autonomous driving, yet recovering reliable 3D geometry from a single image remains challenging due to inherent depth ambiguity. Prior methods project image features into Bird's-Eye-View (BEV) space under a flat-ground assumption, causing geometric distortion on real-world roads. Recent methods instead predict explicit height maps to capture non-planar surfaces, but still rely on sparse anchor-based regression and exploit the recovered geometry merely for spatial transformation rather than semantic understanding. To overcome these limitations, we propose \textbf{HSDF-Lane}, which implicitly models the road surface as a Height-aligned Signed Distance Field (HSDF) over a densely sampled 3D feature volume. Through differentiable rendering, the HSDF jointly produces an accurate height map and surface-aligned features. We further introduce Lane-aware Semantic Positional Encoding (LSPE), which injects a lane-existence prior derived from the surface-aligned features into the transformer queries, coupling geometric structure with semantic guidance. Extensive experiments on the OpenLane benchmark show that HSDF-Lane achieves state-of-the-art performance in both 3D lane detection and height map estimation. The code is available at \url{https://github.com/JiyongBoo/HSDF-Lane}.

\end{abstract}

\section{Introduction}
\label{sec:intro}

3D lane detection is a key perception task for autonomous driving, as it provides essential information about road structure, vehicle localization, and trajectory planning. Monocular camera-based research is particularly attractive due to its cost-effective deployment and simple sensor configuration, while still providing rich appearance cues for identifying lane markings.

Recent advances in deep neural networks have significantly improved visual understanding from monocular images, enabling remarkable progress in 2D lane detection~\cite{pan2018spatial, neven2018towards, tabelini2021keep, tabelini2021polylanenet, feng2022rethinking, ko2021key, liu2021end, zheng2022clrnet, wang2022keypoint, lee2017vpgnet, li2019line, qu2021focus}. However, these approaches inherently lack the geometry information required for reliable 3D applications. Motivated by the need to recover 3D geometry from monocular images for downstream tasks, monocular 3D lane detection has been extensively studied in recent years. Early works primarily adopted Inverse Perspective Mapping (IPM) to transform image features into Bird’s-Eye-View (BEV) representations~\cite{garnett20193d, guo2020gen, pittner20233d, chen2022persformer, li2022reconstruct, ozturk2025glane3d, zheng2024pvalane}. However, BEV transformations relying on flat-ground assumptions often suffer from severe geometric misalignment on non-planar roads. To alleviate this limitation, subsequent methods introduced depth estimation branches or Lift-Splat-Shoot (LSS)~\cite{philion2020lift} mechanisms to lift image features into 3D space~\cite{pittner2024lanecpp, lyu2025depth3dlane, liu2025db3d}. Nevertheless, depth-based lifting remains unstable in far-range regions or rippled surfaces due to the scale ambiguity of monocular depth estimation and the amplification of projection errors during 3D reconstruction. Concurrently, lane query-based methods~\cite{luo2023latr, pittner2025sparselanestp, liu2023petrv2} directly predict 3D lanes from 2D features using 3D positional encoding (PE) to compensate for insufficient geometric cues.

\begin{figure}[t]
  \centering
  \includegraphics[width=1.0\textwidth]{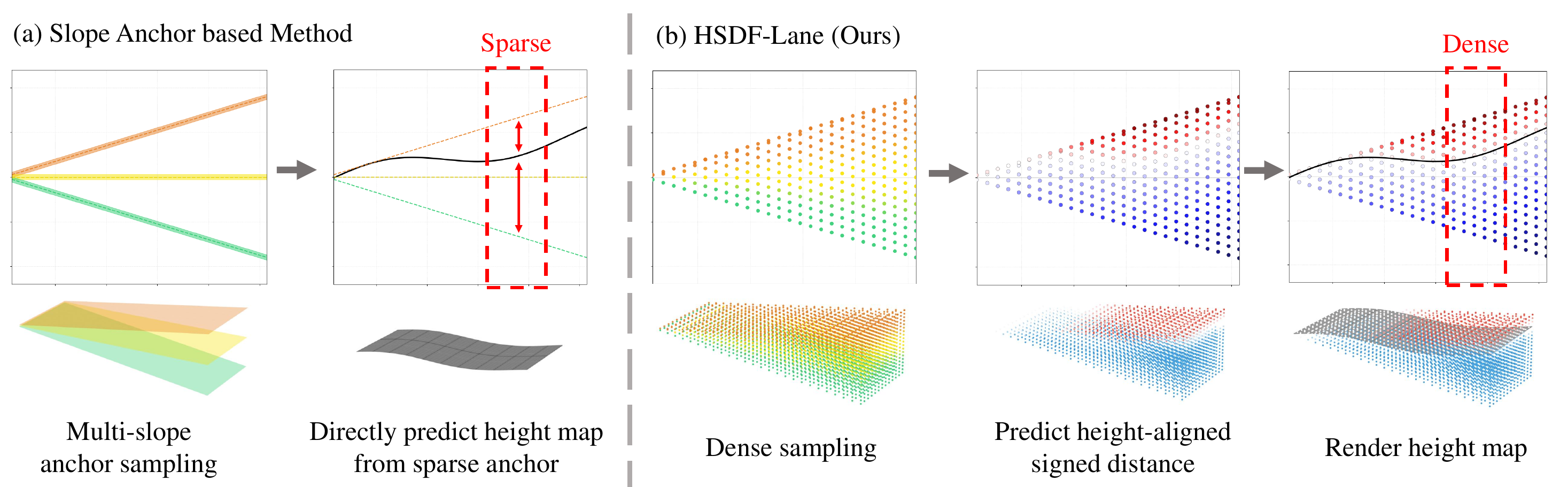}
  \caption{\textbf{Comparison between slope-anchor methods and HSDF-Lane.} (a) Previous approaches~\cite{park2025heightlane,park2025sc} rely on a small number of predefined slope anchors and directly regress the height map, which leads to sparse sampling and potential geometric misalignment. (b) HSDF-Lane densely samples the vertical space and predicts a height-aligned signed distance from each sample, enabling dense supervision and geometrically consistent rendering.}
  \label{fig:teaser}
\end{figure}

Recent approaches, HeightLane~\cite{park2025heightlane} and SC-Lane~\cite{park2025sc}, introduce a height map based architecture for monocular 3D lane detection. Unlike depth, which measures the line-of-sight distance from the camera, a height map represents the absolute vertical coordinate at each ground-plane location on the BEV grid. The height map-based modeling significantly alleviates geometric inconsistencies in BEV space. However, the framework still approximates complex road geometry using a sparse set of predefined planar anchors. Existing slope anchors rely on planar assumptions, which inherently introduce geometric distortion on complex roads. Furthermore, as distance increases, the spatial gaps between these discrete anchors widen significantly, making it inherently difficult to accurately regress the absolute road height, as shown in~\Fref{fig:teaser} (a).

To address the limitations discussed above, we propose HSDF-Lane, a novel 3D lane detection framework that combines dense spatial sampling with implicit surface modeling and effective utilization of spatially aligned features. Rather than relying on sparse slope anchors, HSDF-Lane densely samples the scene along the vertical axis to capture complex road surfaces. The resulting volume is modeled with a Height-aligned Signed Distance Field (HSDF), whose differentiable rendering naturally produces both an explicit height map and spatially consistent features, as illustrated in~\Fref{fig:teaser}(b). To stabilize the HSDF representation and enforce physically consistent surfaces, a height-directional Eikonal loss is incorporated during optimization. Furthermore, to address the limited semantic exploitation in previous approaches, we introduce Lane-aware Semantic Positional Encoding (LSPE). Specifically, a lane heatmap predicted from the rendered HSDF features is converted into a semantic embedding, which then augments the height-based positional encoding to provide semantic guidance for queries. By jointly leveraging dense surface representations and semantic priors, HSDF-Lane enables robust height estimation and accurate 3D lane detection.

\renewcommand{\labelitemi}{$\bullet$}

In summary, the main contributions of this paper are as follows:
\begin{itemize}
    \item We propose a dense 3D sampling framework with a Height-aligned Signed Distance Field (HSDF) representation and differentiable rendering pipeline, which replaces sparse anchor-based height regression to produce surface-aligned features and an accurate height map.
    
    \item We propose Lane-aware Semantic Positional Encoding (LSPE), which incorporates lane existence probabilities derived from rendered features as semantic priors for transformer queries.
    
    \item Extensive experiments on the OpenLane benchmark demonstrate that HSDF-Lane significantly improves both height map estimation and 3D lane detection accuracy, achieving state-of-the-art performance.
\end{itemize}

\section{Related Work}
\subsection{3D Lane Detection}
Motivated by the rapid progress in 2D lane detection~\cite{ghafoorian2018gan, hou2019learning, huval2015empirical, ko2021key, lee2017vpgnet, li2019line, neven2018towards, pan2018spatial, pizzati2019lane, qu2021focus, tabelini2021keep, wang2022keypoint, zou2019robust, xu2020curvelane, zheng2021resa, jin2022eigenlanes, su2021structure, wang2018lanenet, zheng2022clrnet, xu2022rclane, feng2022rethinking, liu2021end, tabelini2021polylanenet}, recent works increasingly explore monocular 3D lane detection to meet the demands of autonomous driving.

Early approaches~\cite{guo2020gen, garnett20193d, li2022reconstruct, pittner20233d} primarily utilized IPM to transform front-view (FV) features into a BEV representation. However, the planar-ground assumption and discretization inherent in IPM often introduce geometric bias and information loss. To mitigate such limitations, subsequent studies~\cite{chen2022persformer, wang2023bev, zheng2024pvalane} incorporate learnable view transformations or feature alignment mechanisms. PersFormer~\cite{chen2022persformer} employs Deformable Cross-Attention (DCA) for spatial feature transformation, while BEV-LaneDet~\cite{wang2023bev} introduces a virtual camera module to improve robustness under data augmentation. PVALane~\cite{zheng2024pvalane} proposes view-agnostic feature alignment to reduce the discrepancy between FV and BEV representations. Other studies~\cite{pittner2024lanecpp, li2024grouplane} adopt Lift-Splat-Shoot (LSS)~\cite{philion2020lift} as an alternative strategy for constructing BEV representations. Recently, GLane3D~\cite{ozturk2025glane3d} achieves strong performance among IPM-based methods by introducing non-uniform BEV sampling and a graph-based lane assembly scheme.

To overcome the geometric misalignment introduced by BEV transformations, anchor-based and query-based methods~\cite{huang2023anchor3dlane, luo2023latr, liu2023petrv2, bai2022curveformer} have been proposed to predict 3D lanes directly from FV images. Recent studies further improve direct 3D lane prediction through lane-aware attention and enhanced lane representations~\cite{pittner2025sparselanestp, chang2025rethinking}. In particular, SparseLaneSTP~\cite{pittner2025sparselanestp} demonstrates notable gains by leveraging spatio-temporal priors and a novel curve representation.

Recognizing the importance of geometric cues, several approaches explicitly incorporate depth estimation. SALAD~\cite{yan2022once} reconstructs 3D lanes by jointly predicting depth and lane segmentation in the image plane. More recently, Depth3DLane~\cite{lyu2025depth3dlane} distills knowledge from foundation depth models~\cite{yang2024depth}, and DB3D-L~\cite{liu2025db3d} constructs BEV features with explicit depth supervision.
Distinct from these depth-dependent approaches, HeightLane~\cite{park2025heightlane} introduces explicit height map estimation with multi-slope anchors and height-guided DCA for BEV transformation, and SC-Lane~\cite{park2025sc} further improves it with slope-aware adaptive features and temporal consistency.

\subsection{Road Height Estimation}

Road height estimation has been studied in several related contexts, including BEV-based perception, aerial elevation mapping, and ground height prediction. (i) \emph{Object-centric height prediction}, exemplified by several existing approaches~\cite{yang2023bevheight, wu2024heightformer}, which estimate height as an auxiliary cue for BEV-based perception and 3D object detection. Recent frameworks, including Height3D~\cite{zhang2025height3d}, HeightAware-BEV~\cite{zhou2025heightaware}, and BEVSpread~\cite{wang2024bevspread}, further predict height distributions to facilitate view transformation and 2D-to-3D mapping. (ii) \emph{Aerial imagery height estimation}, represented by HeightFormer~\cite{chen2024heightformer}, which focuses on large-scale elevation mapping and differs substantially from onboard driving scenarios. (iii) \emph{Ground height estimation}, such as HeightMapNet~\cite{qiu2025heightmapnet}, which leverages multi-camera inputs and map priors but relies on discretized formulations due to limited dense supervision.

HeightLane~\cite{park2025heightlane} and SC-Lane~\cite{park2025sc} extend road height estimation to monocular 3D lane detection by introducing dense BEV height maps for explicit road-surface modeling, while SC-Lane additionally establishes unified metrics for quantitative evaluation. However, both methods remain \emph{anchor-based}, requiring one-shot regression of absolute height from sparse and slope-dependent projections, which becomes particularly challenging on distant or highly non-planar roads.

Meanwhile, implicit field representations~\cite{mildenhall2021nerf} have shown strong capability for continuous 3D modeling~\cite{wang2021neus, yariv2021volume} in driving scenes. SurroundSDF~\cite{liu2024surroundsdf} adopts an SDF-based formulation with Eikonal-style regularization for camera-only surface perception, and InfiniDepth~\cite{yu2026infinidepth} models depth as a neural implicit field enabling continuous, arbitrary-resolution querying. Despite their effectiveness, these approaches primarily target general scene geometry, rather than explicitly optimizing dense road height maps on a BEV grid. Motivated by this gap, our work leverages an HSDF-based implicit formulation tailored to road height estimation for 3D lane detection.

\section{Method}
\label{sec:method}

We define the vehicle coordinate system where the $x$-axis points to the forward (longitudinal) direction and the $y$-axis points to the left (lateral) direction in the BEV. The region of interest is quantized into a regular BEV grid $\mathcal{G}$ of size $H \times W$, where each grid cell $(i, j)$ corresponds to a physical location $(x_i, y_j)$.
Given a single front-view image $\mathcal{I}$, our goal is to predict 3D lanes and a height map on the grid. Specifically, we estimate a dense height map $\mathcal{H} \in \mathbb{R}^{H \times W}$, where each value represents the vertical height $z$~\cite{park2025heightlane}. For 3D lane detection, we adopt the key-point based approach~\cite{wang2023bev} to predict a lane segmentation map $\mathcal{S} \in \mathbb{R}^{H \times W \times 1}$, lateral offsets $\mathcal{O} \in \mathbb{R}^{H \times W \times 2}$, and embeddings $\mathcal{E} \in \mathbb{R}^{H \times W \times C_\text{emb}}$. These dense predictions are clustered to form lane instances, which are subsequently lifted into 3D space using the estimated road surfaces.

\begin{figure}[t]
  \centering
  \includegraphics[width=1.0\textwidth]{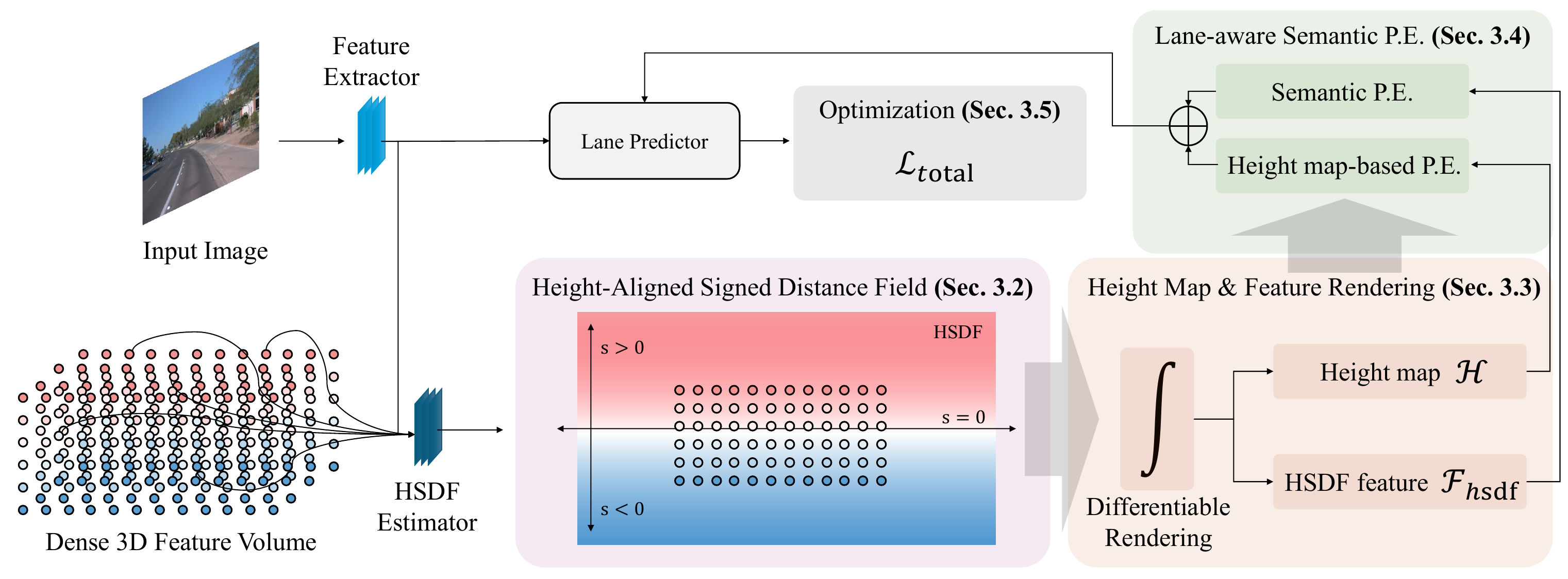}
  \caption{\textbf{Overview of HSDF-Lane.} Given a monocular image, dense 3D features are constructed by vertically sampling the BEV grid. The HSDF estimator predicts a Height-aligned Signed Distance Field (HSDF) (\Sref{sec:method_sampling_hsdf}), and following differentiable rendering (\Sref{sec:method_render}) produces a height map $\mathcal{H}$ and an HSDF-based feature $\mathcal{F}_{hsdf}$. The feature is further enhanced by Lane-aware Semantic Positional Encoding (LSPE) (\Sref{sec:method_lspe}), where a semantic positional encoding derived from a predicted HSDF feature $\mathcal{F}_{hsdf}$ is combined with height-based positional encoding to provide a semantic prior for queries. The refined queries are decoded by the lane predictor, and the entire framework is optimized end-to-end (\Sref{sec:method_loss}).}
  \label{fig:pipeline}
\end{figure}

\subsection{Overview}
\label{sec:method_overview}

\Fref{fig:pipeline} illustrates the overall pipeline of HSDF-Lane. The input front-view image $\mathcal{I}$ is encoded by a backbone to obtain 2D features $\mathcal{F}_{FV}$. For each BEV cell in $\mathcal{G}$, we densely sample points along the vertical $z$-axis and project them onto the image plane using intrinsics $\mathbf{K}$ and extrinsics $\mathbf{T}$ to gather corresponding image features. Stacking the sampled features forms a dense 3D feature volume. An MLP then predicts the \emph{Height-aligned Signed Distance Field (HSDF)}, which represents the vertical signed distance to the road surface. HSDF enables differentiable rendering of both the height map $\mathcal{H}$ and the HSDF-based feature map $\mathcal{F}_{hsdf}$. To incorporate semantic cues from $\mathcal{F}_{hsdf}$ into lane decoding, we introduce Lane-aware Semantic Positional Encoding (LSPE). A BEV lane heatmap $\mathcal{M} \in \mathbb{R}^{H \times W \times 1}$ is predicted from $\mathcal{F}_{hsdf}$ and used to augment BEV queries as an additional positional encoding for the transformer-based lane head. The enriched BEV queries are updated via deformable cross-attention using 3D reference points derived from $\mathcal{H}$. Finally, the updated queries are fed into the 3D lane head to produce the final 3D predictions, ($\mathcal{S}, \mathcal{O}, \mathcal{E}$).

\subsection{Dense 3D Feature Sampling \& HSDF Modeling}
\label{sec:method_sampling_hsdf}

Unlike previous methods that rely on sparse slope anchors, we perform dense sampling along the vertical axis to capture detailed road geometry.

For each grid cell $(x_i, y_j)$, the vertical search range $[z_{\min}, z_{\max}]$ is defined proportional to the longitudinal distance $x_i$:
\begin{equation}
z_{\min, \max} = x_i \cdot \tan(\pm\theta),
\label{eq:z_range}
\end{equation}
where $\theta$ denotes the predefined maximum slope angle. Within this vertical range, we sample $N$ points along the $z$-axis. To achieve dense sampling across all distances by allocating fewer points at near regions and more points at far regions, we introduce an adaptive sampling strategy based on a target \emph{Vertical Sampling Resolution} $\delta_z$. The required number of samples $N$ for a given cell $(x_i, y_j)$ is determined as:

\begin{equation}
N(x_i, y_j) = \max\left(2,\ \left\lceil \frac{z_{\max}-z_{\min}}{\delta_z} \right\rceil + 1\right).
\label{eq:N_adaptive}
\end{equation}
To facilitate the softmax operation during the subsequent differentiable rendering phase (detailed in \Sref{sec:method_render}), we strictly enforce a minimum of two samples ($N \ge 2$) in near-field regions. The adaptive sampling rule in \Eref{eq:N_adaptive} guarantees that the spacing between adjacent samples never exceeds $\delta_z$. Consequently, we obtain a set of discrete 3D sampling points $\mathcal{P}_{sampled} = \{ (x_i, y_j, z_k) \mid i \in [0, H), j \in [0, W), k \in [0, N) \}$, where the $z$-coordinates are uniformly partitioned:
\begin{equation}
z_k = z_{\min} + \frac{k}{N-1}(z_{\max} - z_{\min}), \quad k=0,\dots,N-1.
\label{eq:z_sampling}
\end{equation}

Next, we project the 3D sampling points $\mathcal{P}_{sampled}$ onto the image plane using the camera parameters to retrieve pixel-aligned features from the 2D backbone feature $\mathcal{F}_{FV}$. By stacking the retrieved features along the vertical $z$-axis up to $N$, we construct a dense 3D feature volume $\mathcal{V} \in \mathbb{R}^{H \times W \times N_{max} \times C}$. Invalid sample positions are explicitly handled via a padding mask during the rendering phase.

To implicitly model the road surface geometry within $\mathcal{V}$, we propose the \emph{Height-aligned Signed Distance Field (HSDF)}. Unlike conventional SDF formulations that measure isotropic distances in 3D space, our HSDF is strictly \emph{axis-aligned} to the vertical direction, making it directly interpretable for height estimation. We define the ground-truth HSDF value $s$ at a point $(x, y, z)$ as:
\begin{equation}
s(x,y,z) = z - \mathcal{H}_{GT}(x,y),
\label{eq:hsdf_def}
\end{equation}
where $\mathcal{H}_{GT}(x,y)$ is the ground truth road height. A value of $s=0$ indicates the exact road surface, while the sign represents the relative vertical position. Finally, we employ an MLP with a skip connection to predict the HSDF value $\hat{s}$ using the feature volume $\mathcal{V}$:
\begin{equation}
\hat{s}(x_i,y_j,z_k) = \text{MLP}\left( \text{cat}\big[ \mathcal{V}(x_i,y_j,z_k),\ \gamma(x_i,y_j,z_k) \big] \right),
\label{eq:hsdf_prediction}
\end{equation}
where $\gamma(\cdot)$ represents the positional encoding~\cite{mildenhall2021nerf}.

\subsection{HSDF-based Differentiable Height Map \& Feature Rendering}
\label{sec:method_render}

\begin{figure}[t]
  \centering
  \includegraphics[width=0.8\textwidth]{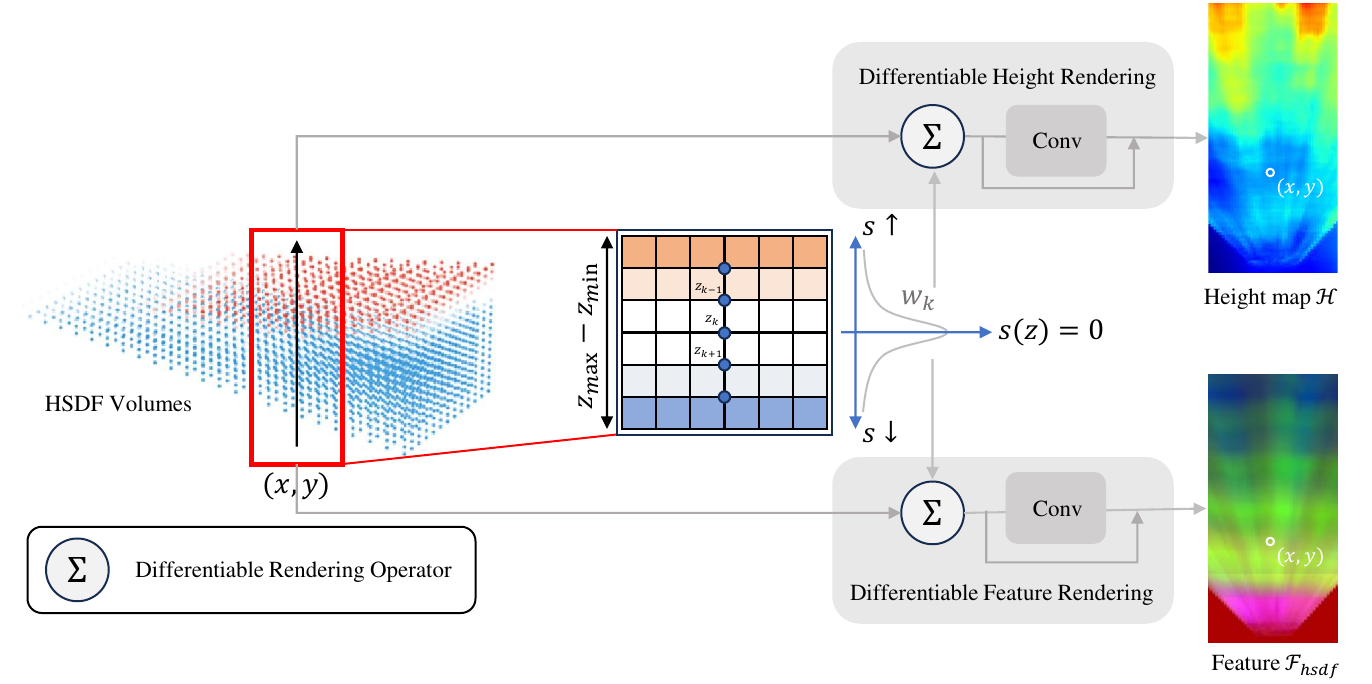}
  \caption{\textbf{HSDF-based differentiable rendering.} For each BEV location $(x,y)$, probability weights $w_k$ are computed from the predicted HSDF values using a softmax operator. These weights are used to aggregate vertical samples along the ray, producing the rendered height map $\mathcal{H}$ and surface-aligned features $\mathcal{F}_{hsdf}$.}
  \label{fig:contrib0001}
\end{figure}

In our framework, the road surface is implicitly represented by the coordinates where the HSDF value is zero, $s(x, y, z) = 0$. To extract the explicit height map from the HSDF representation, we employ a differentiable soft argmin operation. For each grid cell $(x_i, y_j)$, we compute probability weights $w_k$ by applying a softmax function over the valid sampled indices $k$:
\begin{equation}
w_k(x_i, y_j) = \text{softmax}_k\left(-\frac{|\hat{s}(x_i, y_j, z_k)|}{\tau(x_i)}\right).
\label{eq:weights}
\end{equation}
Invalid indices where the sample exceeds the upper boundary ($z_k > z_{\max}$) are explicitly masked out prior to the softmax operation. Here, the temperature $\tau(x_i)$ controls the sharpness of the weight distribution. To ensure consistent optimization behavior across varying vertical search ranges, we scale $\tau$ adaptively:
\begin{equation}
\tau(x_i) = \tau_0 \cdot \text{clip}\left(\frac{z_{\max} - z_{\min}}{\delta_z},\ 0.3,\ 3.0\right),
\label{eq:tau}
\end{equation}
where $\tau_0$ is a learnable base temperature, and $\delta_z$ is the fixed vertical sampling interval defined in \Sref{sec:method_sampling_hsdf}. The clipping operation bounds the scaling factor to $[0.3, 3.0]$, preventing numerical instability from near-zero temperatures while avoiding excessive smoothing that would result from large temperatures.

Using the weights $w_k(x_i, y_j)$, we first compute a preliminary continuous height map $\tilde{\mathcal{H}}$. Instead of simply aggregating the discrete sample coordinates $z_k$, we treat the predicted HSDF value $\hat{s}$ as a local residual to refine the coarse sampling. Specifically, each sample $z_k$ is shifted toward the surface using the predicted distance $\hat{s}$, and the refined coordinates are aggregated via a weighted sum:
\begin{equation}
\tilde{\mathcal{H}}(x_i, y_j) = \sum_{k=0}^{N-1} w_k(x_i, y_j) \cdot \left( z_k - \hat{s}(x_i, y_j, z_k) \right).
\label{eq:height_raw}
\end{equation}
Because this ray-wise rendering operates independently for each grid cell, it may introduce local spatial discontinuities. To improve spatial smoothness while maintaining stable optimization in the early training stages, we apply a zero-initialized $3\times3$ residual convolution over $\tilde{\mathcal{H}}$ to obtain the final height map $\mathcal{H}$.

Concurrently, to extract semantic features corresponding to the road surface for lane understanding, we compute a preliminary feature map $\tilde{\mathcal{F}}_{hsdf}$ by aggregating the feature volume $\mathcal{V}$ using the identical weights:
\begin{align}
\tilde{\mathcal{F}}_{hsdf}(x_i, y_j) = \sum_{k=0}^{N-1} w_k(x_i, y_j) \cdot \mathcal{V}(x_i, y_j, z_k).
\label{eq:feat_render}
\end{align}
To further improve spatial coherence, the aggregated features $\tilde{\mathcal{F}}_{hsdf}$ are refined using a zero-initialized depthwise-pointwise residual block, producing the final HSDF-based feature map $\mathcal{F}_{hsdf}$. The overview of HSDF is illustrated in~\Fref{fig:contrib0001}.

Furthermore, we introduce a \emph{Height-directional Eikonal Regularization}. A property of an SDF is that the magnitude of its gradient is equal to 1, i.e., $\|\nabla f\| = 1$~\cite{gropp2020implicit, wang2021neus, yariv2021volume}. While conventional approaches enforce the constraint on the full 3D gradient, our HSDF is designed specifically for height estimation along the vertical direction. Therefore, the regularization constrains only the partial derivative along the vertical axis, i.e., $\partial s / \partial z = 1$. Because the representation is defined on discrete samples, the derivative is approximated using finite differences:
\begin{equation}
\mathcal{L}_{\text{eik}} = \sum_{(x_i,y_j,z_k) } \left| \frac{\hat{s}(x_i,y_j,z_{k+1})-\hat{s}(x_i,y_j,z_k)} {z_{k+1}-z_k} -1 \right|.
\label{eq:eikonal}
\end{equation}
 This regularization encourages the network to learn a physically meaningful distance field rather than arbitrary scalar values, stabilizing zero-crossing localization and improving height rendering quality.

\subsection{Lane-aware Semantic Positional Encoding (LSPE)}
\label{sec:method_lspe}

\begin{figure}[t]
  \centering
  \includegraphics[width=1.0\textwidth]{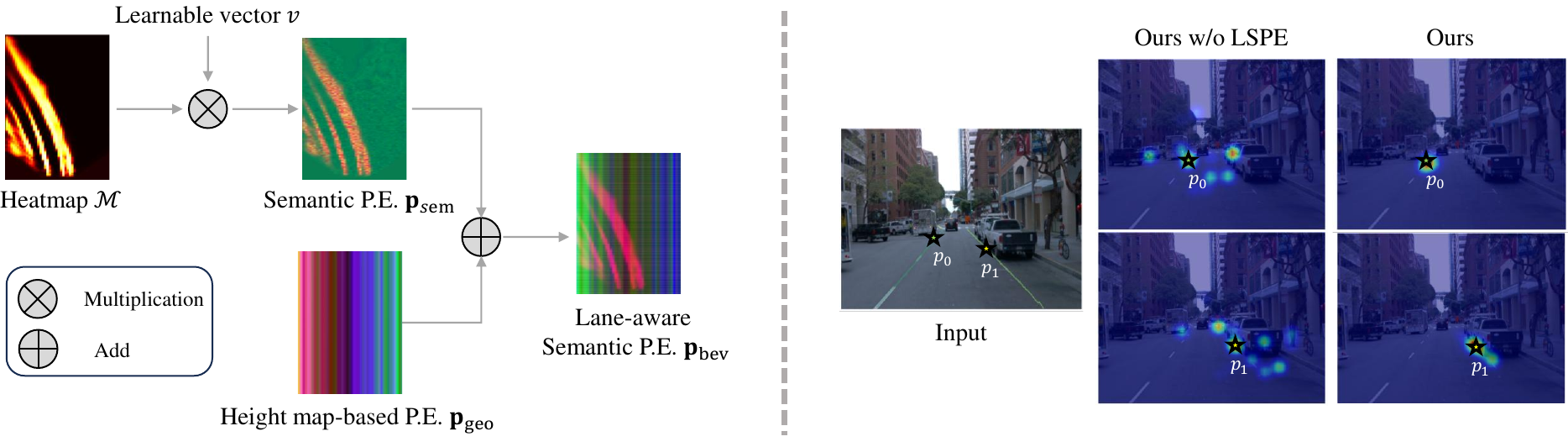}
  \caption{\textbf{(Left) Lane-aware Semantic Positional Encoding.} A Lane heatmap $\mathcal{M}$ is multiplied by learnable vector $\mathbf{v}$ and builds a semantic embedding $\mathbf{p}_{sem}$ that is fused with height map–based positional encoding $\mathbf{p}_{geo}$ to produce the final Lane-aware Semantic positional embedding $\mathbf{p}_{bev}$. \textbf{(Right) Visualization of Effect of LSPE.} We sampled two reference points $p_0$ and $p_1$, and visualized weight of attention. With LSPE, our model shows robust sampling ability to align attention with the lane to enhance lane understanding.}
  \label{fig:contrib0002}
\end{figure}

HSDF-based rendering produces a geometrically aligned feature map $\mathcal{F}_{hsdf}$, which provides a representation for extracting semantic priors. A lane existence heatmap $\mathcal{M} \in \mathbb{R}^{H \times W \times 1}$ is predicted directly from $\mathcal{F}_{hsdf}$ using a lightweight CNN. The auxiliary prediction task captures semantic lane information and encourages the features to concentrate on lane regions, thereby implicitly improving the spatial consistency of $\mathcal{F}_{hsdf}$.

As illustrated in the left panel of \cref{fig:contrib0002}, the predicted heatmap $\mathcal{M}$ is converted into a positional encoding to incorporate the semantic prior into the transformer. Instead of adopting a full learnable embedding, which may introduce noise and interfere with the geometric positional embedding $\mathbf{p}_{\text{geo}}$~\cite{park2025heightlane}, a single learnable vector $\mathbf{v} \in \mathbb{R}^{C}$ is employed. The semantic encoding is obtained by modulating the vector $\mathbf{v}$ with the predicted heatmap probability $\mathcal{M}$. The resulting semantic embedding $\mathbf{p}_{\text{sem}}$ is combined with the geometric positional embedding $\mathbf{p}_{\text{geo}}$  derived from the heightmap $\mathcal{H}$ to construct the final Lane-aware Semantic Positional Encoding $\mathbf{p}_{\text{bev}}$ :
\begin{align}
\mathbf{p}_{\text{sem}}(x_i, y_j) &= \mathcal{M}(x_i, y_j) \cdot \mathbf{v}, \\
\mathbf{p}_{\text{bev}}(x_i, y_j) &= \mathbf{p}_{\text{geo}}(x_i, y_j) + \mathbf{p}_{\text{sem}}(x_i, y_j).
\label{eq:semantic_pe}
\end{align}
The composite embedding $\mathbf{p}_{\text{bev}}$ provides the transformer with a semantic prior indicating regions where lane existence is highly probable, while preserving the geometric structure encoded in $\mathbf{p}_{\text{geo}}$.

Finally, following the height-guided spatial transform framework~\cite{park2025heightlane}, the rendered height map $\mathcal{H}$ and the composite positional embedding $\mathbf{p}_{\text{bev}}$ are used to generate refined BEV features. The refined features are subsequently processed by the lane detection head to estimate 3D lanes.

\subsection{Training Loss}
\label{sec:method_loss}

The training objective of HSDF-Lane is formulated as a weighted sum of all losses:
\begin{equation}
\begin{aligned}
\mathcal{L}_{total} = 
\lambda_{\text{seg}}\mathcal{L}_{\text{seg}}
+  &\lambda_{\text{off}}\mathcal{L}_{\text{off}} 
+\lambda_{\text{emb}}\mathcal{L}_{\text{emb}}\\
+\lambda_{\text{2d}}\mathcal{L}_{\text{2d}}  + 
& \lambda_{\text{geo}}(\mathcal{L}_{\mathcal{H}} + \lambda_{\text{eik}}\mathcal{L}_{\text{eik}})
+\lambda_{\text{hm}}\mathcal{L}_{\text{hm}}.
\label{eq:overall_loss}
\end{aligned}
\end{equation}

Following~\cite{park2025heightlane, wang2023bev}, the segmentation loss $\mathcal{L}_{\text{seg}}$ is implemented as a combination of Binary Cross Entropy (BCE) and IoU losses. The offset loss $\mathcal{L}_{\text{off}}$ is optimized using a BCE loss, while the embedding loss $\mathcal{L}_{\text{emb}}$ employs the discriminative push–pull loss. The auxiliary 2D supervision loss is denoted as $\mathcal{L}_{\text{2d}}$. Geometry-related losses $\mathcal{L}_{\mathcal{H}}$, $\mathcal{L}_{\text{eik}}$, and the lane heatmap loss $\mathcal{L}_{\text{hm}}$ are employed for geometry learning. Detailed formulations of the losses and the weighting coefficients $\lambda$ are provided in the supplementary material.

\section{Experiments}
\label{sec:experiments}

\begin{table}[t]
  \caption{Quantitative comparison on OpenLane Benchmark~\cite{chen2022persformer}. The best results are highlighted in \textbf{bold}, while the second-best results are \underline{underlined}. (${\dagger}$): HSDF-Lane equipped with FPN.
  }
  \label{tab:lane_detection_results}
  \centering
  \resizebox{\columnwidth}{!}{%
  \begin{tabular}{l>{\hspace*{1.5em}}lccccc}
    \toprule
    \multirow{2}{*}{Method} & \multirow{2}{*}{Backbone} & \multirow{2}{*}{F1-Score(\%)$\uparrow$} & \multicolumn{2}{c}{X-error (m) $\downarrow$} & \multicolumn{2}{c}{Z-error (m) $\downarrow$} \\
     & & & \makebox[1.2cm]{near} & \makebox[1.2cm]{far} & \makebox[1.2cm]{near} & \makebox[1.2cm]{far} \\
     
    \midrule
   
    PersFormer\cite{chen2022persformer}        & EfficientNet-B7 & 50.5           & 0.485          & 0.553          & 0.364          & 0.431 \\
   
    BEV-LaneDet\cite{wang2023bev}      & ResNet-34       & 58.4           & 0.309          & 0.659          & 0.244          & 0.631 \\
   
    LATR\cite{luo2023latr}             & ResNet-50       & 61.9           & 0.219 & 0.259          & 0.075 & 0.104 \\
    
    PVALane\cite{zheng2024pvalane}         & Swin-B          & 63.4           & 0.226          & 0.257 & 0.093          & 0.119 \\
    GroupLane\cite{li2024grouplane}        & ConvNext-Base   & 64.1 & 0.320          & 0.441          & 0.233          & 0.402 \\
    HeightLane\cite{park2025heightlane}  & ResNet-50       & 62.7  & 0.240 & 0.266 & 0.116 & 0.165 \\
    SC-Lane\cite{park2025sc}  & ResNet-50       & 64.3  & 0.227 & 0.251 & 0.088 & 0.128 \\
    Rethinking\cite{chang2025rethinking}  & ResNet-50       & 64.7  & 0.205 & 0.255 & 0.074 & 0.105 \\
    GLane3D-B\cite{ozturk2025glane3d}  & ResNet-50    &   63.9  & \underline{0.193}  &  0.234  &  \textbf{0.065}  & \textbf{0.090} \\
    
    SparseLaneSTP\cite{pittner2025sparselanestp}  & ResNet-50       & 66.1  & 0.203 & 0.240 & \underline{0.066} & \underline{0.092} \\
    \rowcolor{gray!20}\textbf{ HSDF-Lane (Ours)}  & ResNet-50       & \underline{66.3}  & 0.201 & \textbf{0.223} & 0.088 & 0.114 \\
    \rowcolor{gray!20}\textbf{ HSDF-Lane$^{\dagger}$ (Ours)}  & ResNet-50       & \textbf{66.9}  & \textbf{0.186} & \underline{0.226} & 0.084 & 0.114 \\
  \specialrule{\heavyrulewidth}{0pt}{0pt}
  \end{tabular}%
  }
\end{table}
\begin{table}[t]
  \caption{Quantitative results comparison by scenario on the OpenLane~\cite{chen2022persformer}  validation set using F1-score (\%). The best results for each scenario are highlighted in \textbf{bold} and second-best results are \underline{underlined}. (${\dagger}$): HSDF-Lane equipped with FPN.}
  \label{tab:openlane_results}
  \centering
  \setlength{\tabcolsep}{8pt} 
  \resizebox{\columnwidth}{!}{%
  \begin{tabular}{l|ccccccc}
    \toprule
    \textbf{Method} & \textbf{All} & \makecell{\textbf{Up \&} \\ \textbf{Down}} & \textbf{Curve} & \makecell{\textbf{Extreme} \\ \textbf{Weather}} & \textbf{Night} & \textbf{Intersection} & \makecell{\textbf{Merge} \\ \textbf{\& Split}} \\
    \midrule
    BEVLaneDet\cite{wang2023bev} & 58.4 & 48.7 & 63.1 & 53.4 & 53.4 & 50.3 & 53.7 \\
    LATR\cite{luo2023latr}       & 61.9 & 55.2 & 68.2 & 57.1 & 55.4 & 52.3 & 61.5 \\
    HeightLane\cite{park2025heightlane} & 62.7 & 53.6 & 69.3 & 55.4 & 54.6 & 54.1 & 61.1 \\
    PVALane\cite{zheng2024pvalane}    & 62.7 & 54.1 & 67.3 & \textbf{62.0} & 57.2 & 53.4 & 60.0 \\
    SC-Lane\cite{park2025sc} & 64.3 & 54.6 & 70.7 & 58.1 & 56.5 & 56.1 & 63.0 \\
    GLane3D-B\cite{ozturk2025glane3d} & 63.9 & \underline{58.2} & 71.1 & 60.1 & \textbf{60.2} & 55.0 & 64.8 \\
    SparseLaneSTP\cite{pittner2025sparselanestp} & 66.1 & 57.3 & 73.0 & 60.1 & 58.3 & 58.2 & \underline{66.5} \\
    \rowcolor{gray!20}\textbf{HSDF-Lane~(Ours)} & \underline{66.3} & \textbf{58.9} & \underline{73.7} & \underline{60.7}  & 57.4 & \underline{58.3} & 65.7 \\
    \rowcolor{gray!20}\textbf{HSDF-Lane$^{\dagger}$~(Ours)} & \textbf{66.9} & \underline{58.2} & \textbf{74.2} & 60.5  & \underline{59.6} & \textbf{59.2} & \textbf{66.8} \\
  \specialrule{\heavyrulewidth}{0pt}{0pt}
  \end{tabular}%
  }
\end{table}

\subsection{Experimental Setup}

\noindent \textbf{Datasets.} 
To validate our framework, we conducted experiments on the extensive OpenLane benchmark~\cite{chen2022persformer}, which is derived from the Waymo Open Dataset~\cite{sun2020scalability}. The OpenLane benchmark encompasses 1,000 driving scenes, totaling 200,000 frames split into 150,000 for training and 50,000 for validation, alongside 880,000 spatial lane annotations in 3D coordinates. The evaluation set is categorized into six distinct scenarios: Up \& Down, Curve, Extreme Weather, Night, Intersection, and Merge \& Split. Furthermore, ablation experiments are carried out on OpenLane-300, a representative subset consisting of 300 scenes. 
For height map training, we directly utilize the ground-truth dense height maps provided by HeightLane~\cite{park2025heightlane}, which were generated by accumulating Waymo LiDAR point clouds. These height maps are structured in a $200 \times 48$ BEV grid format, covering a physical range of $x \in [3\text{m}, 103\text{m}]$ and $y \in [-12\text{m}, 12\text{m}]$ with a spatial resolution of 0.5m per cell.

\noindent \textbf{Baselines.}
We compare our method with several strong monocular 3D lane detection baselines, including PersFormer~\cite{chen2022persformer}, BEV-LaneDet~\cite{wang2023bev}, LATR~\cite{luo2023latr}, PVALane~\cite{zheng2024pvalane}, GroupLane~\cite{li2024grouplane}, and HeightLane~\cite{park2025heightlane}, as well as recent state-of-the-art methods such as SC-Lane~\cite{park2025sc}, Rethinking~\cite{chang2025rethinking}, GLane3D-B~\cite{ozturk2025glane3d}, and SparseLaneSTP~\cite{pittner2025sparselanestp}. For height estimation, we follow the evaluation protocol of SC-Lane and report comparisons with HeightLane and SC-Lane. For qualitative comparisons, we visualize results from HSDF-Lane, HeightLane, and LATR. Other recent methods, such as GLane3D-B and SparseLaneSTP, are not included in the qualitative comparison due to the lack of publicly available code.

\noindent \textbf{Evaluation Metrics.}
We utilized the established metrics from Gen-LaneNet~\cite{guo2020gen} to assess 3D lane detection performance, reporting the F1-score alongside positional errors along the X and Z axes. Here, the X- and Z-errors denote lateral and vertical deviations, differing from the vehicle coordinate system in~\Sref{sec:method}. Specifically, a predicted lane is treated as a true positive only when a minimum of 75\% of its constituent points fall within a 1.5-meter distance threshold from the corresponding ground truth. Spatial deviations (X and Z errors) are computed over two distinct longitudinal intervals: the near range (0 to 40 meters) and the far range (40 to 100 meters). For height map evaluation, we follow the metrics proposed in SC-Lane~\cite{park2025sc}. Specifically, we report Mean Absolute Error (MAE), Root Mean Square Error (RMSE), and threshold-based accuracies at 0.05m, 0.1m, and 0.2m.



\subsection{Implementation Details}
\noindent \textbf{Network Architecture.}
We use ResNet-50~\cite{he2016deep} as the backbone network to extract image features $\mathcal{F}_{FV}$ with a stride of 16 and a channel size of 1024. The input images are resized to $600 \times 800$. We adopt the transformer decoder and 3D lane detection head from HeightLane~\cite{park2025heightlane}. To ensure computational efficiency, the transformer decoder is configured with only two layers. We employ four attention heads, with four sampling points per head for the Deformable Cross-Attention (DCA).

We additionally report HSDF-Lane$^{\dagger}$, a variant that adopts a Feature Pyramid Network (FPN)~\cite{lin2017feature} following LATR~\cite{luo2023latr}. Instead of relying on a single-scale feature, the FPN aggregates multi-scale features from the backbone at strides of $8$, $16$, and $32$, each with a channel size of $256$. The resulting multi-scale features are jointly fed into the transformer decoder, which is configured with four layers for this variant. All other settings remain identical to the base model.

\noindent \textbf{Spatial and SDF Settings.}
The BEV space is quantized into a $200 \times 48$ grid, representing a physical range of $x \in [3\text{m}, 103\text{m}]$ in the longitudinal direction and $y \in [-12\text{m}, 12\text{m}]$ in the lateral direction, which yields a grid resolution of 0.5m per cell. For the dense 3D sampling, the predefined maximum slope angle used to define the vertical search range $[z_{\min}, z_{\max}]$ is set to $\theta = 5^\circ$. We set the target vertical sampling resolution $\delta_z$ to $1.5\text{m}$. Accordingly, the maximum number of sample points along the $z$-axis is bounded by $N_{\max} = 13$. The learnable temperature for the soft argmin rendering operation is initialized to $\tau_0 = 0.3$.

\begin{figure}[t]
  \centering
  \includegraphics[width=1.0\textwidth]{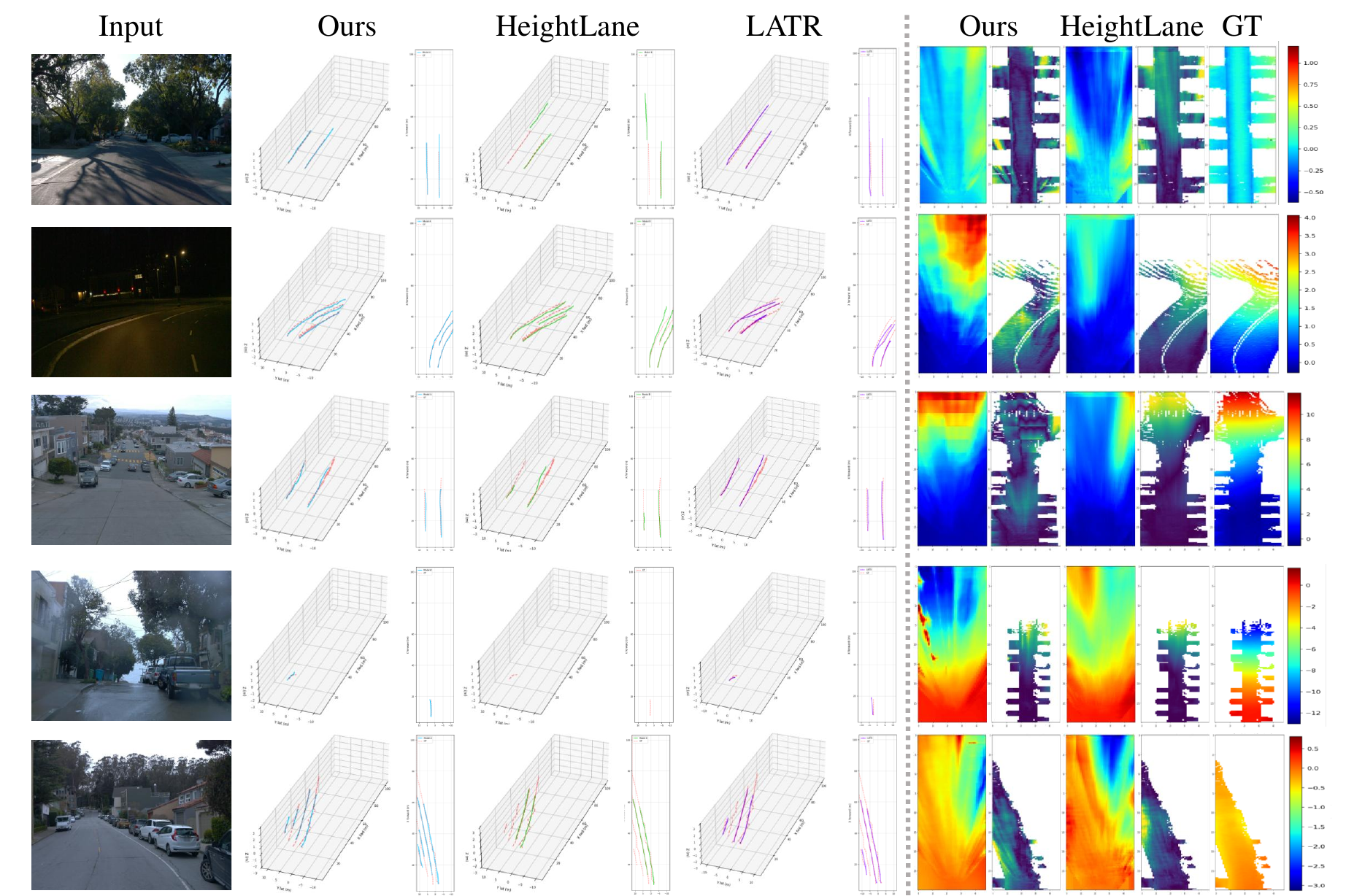} 
  \caption{\textbf{Qualitative comparison of 3D lane detection and road height estimation.} From left to right: input image $\mathcal{I}$, and 3D lane predictions of \textcolor{blue!60}{HSDF-Lane (ours)}, \textcolor{green}{HeightLane}~\cite{park2025heightlane}, and \textcolor{purple!60}{LATR}~\cite{luo2023latr}. The \textcolor{red}{ground truth (GT)} is indicated by red dashed lines. The right panel compares the estimated height maps $\mathcal{H}$ and their corresponding error maps (masked by valid GT regions) for our method and HeightLane, alongside the GT height map $\mathcal{H}_\text{GT}$. Other recent methods~\cite{ozturk2025glane3d,pittner2025sparselanestp} lack public code and are thus excluded. More qualitative results are in the supplementary material.}
  \label{fig:qual}
\end{figure}

\subsection{Quantitative Results}
\Tref{tab:lane_detection_results} presents the quantitative comparison with existing monocular 3D lane detection methods on the OpenLane~\cite{chen2022persformer} validation set. HSDF-Lane achieves the best F1-score among existing methods, and its FPN-equipped variant HSDF-Lane$^{\dagger}$ further improves it, yielding the top two results and outperforming both anchor-based approaches such as HeightLane~\cite{park2025heightlane} and SC-Lane~\cite{park2025sc} and query-based methods including SparseLaneSTP~\cite{pittner2025sparselanestp}. Both models also achieve competitive geometric accuracy, with HSDF-Lane obtaining the best far-range X-error and HSDF-Lane$^{\dagger}$ the best near-range X-error. We note that Z-error is influenced by the representation used for lane estimation. Height-map-based methods~\cite{park2025heightlane, park2025sc}, including our HSDF-Lane, predict and supervise the height over the entire BEV grid. In contrast, other recent methods such as LATR~\cite{luo2023latr}, SparseLaneSTP~\cite{pittner2025sparselanestp}, and GLane3D~\cite{ozturk2025glane3d} directly regress the height at each lane point and apply the Z-loss only at those points. Because the supervision is concentrated on lane locations, such methods tend to report lower Z-error, so the values are not always directly comparable across methods. Nevertheless, HSDF-Lane remains competitive through the proposed HSDF formulation.

To analyze performance under different environmental conditions, \Tref{tab:openlane_results} reports F1-score across the scenario-based subsets of OpenLane. HSDF-Lane performs best in the \emph{Up \& Down} scenario, while HSDF-Lane$^{\dagger}$ attains the best results in \emph{Curve}, \emph{Intersection}, and \emph{Merge \& Split}, as well as the best overall F1-score. These results demonstrate that the proposed HSDF-based representation and LSPE effectively improve lane detection under complex road geometries and diverse environments.

\begin{table}[ht]
\centering
\caption{Comparison of height estimation performance. The best results are highlighted in \textbf{bold} and second-best results are \underline{underlined}. (${\dagger}$): HSDF-Lane equipped with FPN.}
\label{tab:height_estimation}
\newcolumntype{Y}{>{\centering\arraybackslash}X} 
\begin{tabularx}{0.8\linewidth}{c|YY|YYY}
\hline
\multirow{2}{*}{\textbf{Method}} & \multicolumn{2}{c|}{\textbf{Error Metrics} ($\downarrow$)} & \multicolumn{3}{c}{\textbf{Accuracy Metrics} ($\uparrow$)} \\ \cline{2-6} 
 & \textbf{MAE} & \textbf{RMSE} & \textbf{@0.05} & \textbf{@0.1} & \textbf{@0.2} \\ \hline
HeightLane~\cite{park2025heightlane} & 0.235 & 0.343 & 0.253 & 0.442 & 0.673 \\
SC-Lane~\cite{park2025sc} & 0.176 & \textbf{0.259} & 0.293 & 0.507 & 0.756 \\ 
\rowcolor{gray!20}\textbf{HSDF-Lane} & \underline{0.158} & 0.313 & \underline{0.294} & \underline{0.512} & \underline{0.763} \\ 
\rowcolor{gray!20}\textbf{HSDF-Lane$^{\dagger}$} & \textbf{0.149} & \underline{0.272} & \textbf{0.307} & \textbf{0.527} & \textbf{0.776} \\\hline
\end{tabularx}
\end{table}

\Tref{tab:height_estimation} presents the quantitative comparison of road height estimation. Both variants outperform HeightLane and SC-Lane across most metrics, with HSDF-Lane$^{\dagger}$ obtaining the lowest MAE and the highest threshold-based accuracies. The only exception is RMSE, where HSDF-Lane is slightly higher than SC-Lane. The anchor-based regression and temporal consistency loss adopted in SC-Lane suppress outliers and yield lower RMSE, but constrain the representable height range. In contrast, the HSDF-based formulation captures a substantially wider height range, resulting in a marginally higher RMSE. Overall, these results confirm that the proposed HSDF-based representation yields more accurate height estimation than previous anchor-based methods.

\subsection{Qualitative Results}
\Fref{fig:qual} presents qualitative comparisons with representative monocular 3D lane detection methods on the OpenLane dataset~\cite{chen2022persformer}. Compared with HeightLane~\cite{park2025heightlane} and LATR~\cite{luo2023latr}, HSDF-Lane produces more geometrically consistent lane predictions, particularly in challenging scenarios such as curved roads, nighttime conditions, and complex urban environments. In addition to improved lane detection, the proposed method also provides accurate road height estimation. As shown in the right panel of \Fref{fig:qual}, the predicted height maps of HSDF-Lane exhibit smoother and more consistent surface structures, while the corresponding error maps indicate smaller deviations from the ground truth compared with HeightLane. These results demonstrate that modeling the road surface using the proposed HSDF representation enables more reliable geometric understanding, which further benefits downstream 3D lane detection.

\subsection{Ablation Studies}
To evaluate the contributions of the proposed components of HSDF-Lane, we conduct ablation studies on the OpenLane-300 subset~\cite{chen2022persformer}. The quantitative results are reported in~\Tref{tab:ablation}.

\vspace{2mm}
\noindent \textbf{Effect of Explicit Height Map Supervision.} 
Using only the Eikonal regularization (Exp.~A) fails to capture accurate road elevations, resulting in poor lane detection performance. Introducing explicit height map supervision $\mathcal{L}_{\mathcal{H}}$ (Exp.~B) substantially reduces MAE and RMSE while significantly improving threshold-based accuracies. This improved geometric alignment leads to a higher F1-score, confirming that direct supervision on the rendered height map is essential for learning the correct road topology.

\vspace{2mm}
\noindent \textbf{Effect of Height-directional Eikonal Regularization.}
Comparing Exp.~B and Exp.~C shows that adding the height-directional Eikonal regularization $\mathcal{L}_{\text{eik}}$ further improves the stability of HSDF-based height estimation. In particular, $\mathcal{L}_{\text{eik}}$ substantially reduces RMSE and slightly lowers MAE, indicating fewer extreme height deviations and a more physically consistent vertical distance field. While the threshold-based accuracies change only marginally, the improved global consistency leads to a gain in F1-score.

\vspace{2mm}
\noindent \textbf{Effect of LSPE.}
Comparing Exp.~C and Exp.~D demonstrates the benefit of incorporating LSPE, which introduces the heatmap supervision $\mathcal{L}_\text{hm}$. LSPE enhances the F1-score and achieves the best threshold-based accuracies, suggesting that the lane heatmap prior helps the model focus on lane-relevant regions and improve height estimation in a way that is more aligned with lane perception. Although RMSE increases, the overall gains in detection accuracy and threshold-based height metrics indicate that LSPE provides complementary semantic guidance beyond purely geometric supervision. This observation highlights that semantic guidance reinforces geometric supervision and plays a key role in improving 3D lane detection performance.

\begin{table}[h]
\centering
\caption{Ablation study on the key components of HSDF-Lane evaluated on the OpenLane-300 subset~\cite{chen2022persformer}. \textbf{Bold} and \underline{underlined} values indicate the best and second-best performances.}
\label{tab:ablation}
\newcolumntype{Y}{>{\centering\arraybackslash\hsize=0.3\hsize}X}
\newcolumntype{Z}{>{\centering\arraybackslash\hsize=0.5\hsize}X}

\begin{tabular}{c|ccc|c|cc|ccc}
\Xhline{3\arrayrulewidth} 
\multirow{2}{*}{Exp.} & \multicolumn{3}{c|}{Preserve ($\checkmark$)} & \multirow{2}{*}{F1-score ($\uparrow$)} & \multicolumn{2}{c|}{Error ($\downarrow$)} & \multicolumn{3}{c}{Accuracy ($\uparrow$)} \\
\cline{2-4} \cline{6-10}
& $\mathcal{L}_{\text{eik}}$ & $\mathcal{L}_{\mathcal{H}}$ & LSPE &  & MAE & RMSE & Acc@0.05 & Acc@0.1 & Acc@0.2\\ \hline
A & $\checkmark$  &  &  & 71.8 & 0.512 & 0.648 & 0.052 & 0.105 & 0.197 \\
B & & $\checkmark$ & & 74.2 & 0.161 & 0.333 & \underline{0.283} & \underline{0.498} & \underline{0.755}\\
C & $\checkmark$ & $\checkmark$ &  & \underline{74.7} & \textbf{0.157} & \textbf{0.286} & 0.275 & 0.487 & 0.750 \\
D & $\checkmark$ & $\checkmark$ & $\checkmark$ & \textbf{75.6} & \underline{0.158} & \underline{0.326} & \textbf{0.288} & \textbf{0.503} & \textbf{0.767}  \\ 
\Xhline{3\arrayrulewidth} 
\end{tabular}
\end{table}

\section{Conclusion}

In this paper, we presented HSDF-Lane, a novel monocular 3D lane detection framework that addresses the limitations of sparse anchor-based height regression and insufficient semantic utilization. To capture complex road geometry, we replaced predefined slope anchors with a dense feature volume modeled by a Height-aligned Signed Distance Field (HSDF). Stabilized by a Height-directional Eikonal regularization, this implicit formulation enables accurate differentiable rendering of a dense height map and surface-aligned features. We further proposed Lane-aware Semantic Positional Encoding (LSPE), which predicts a lane existence heatmap from the surface-aligned features and embeds it into the transformer queries to integrate semantic priors with the rendered geometry. Extensive experiments on the OpenLane benchmark demonstrate that HSDF-Lane achieves state-of-the-art performance in monocular 3D lane detection while improving road height estimation accuracy.

\noindent\textbf{Limitations.} Despite these results, several limitations remain. The height map is predicted over a fixed BEV grid without identifying the drivable road surface, training requires dense LiDAR-derived height-map supervision, and experiments are limited to a monocular setting without multi-view, temporal, or multi-modal extensions. Overcoming these limitations remains future work.


\section*{Acknowledgements}
This research was supported by the Challengeable Future Defense Technology Research and Development Program through the Agency For Defense Development~(ADD) funded by the Defense Acquisition Program Administration~(DAPA) in 2026~(No.915102201).

%
%
\bibliographystyle{splncs04}
\bibliography{main}\


\setcounter{section}{0}
\setcounter{figure}{0}
\setcounter{table}{0}
\setcounter{equation}{0}

\renewcommand{\thesection}{S\arabic{section}}
\renewcommand{\thefigure}{S\arabic{figure}}
\renewcommand{\thetable}{S\arabic{table}}
\renewcommand{\theequation}{S\arabic{equation}}

\title{Supplementary Material for \\ HSDF-Lane: Height-Aligned Signed Distance Field with Semantic Lane Prior for 3D Lane Detection} 
\titlerunning{HSDF-Lane}

\author{Jiyong Boo\orcidlink{0009-0000-6129-635X} \and
Byeongin Joung\orcidlink{0009-0004-0321-0430} \and
Hyemin Yang \orcidlink{0009-0002-2112-7804} \and
Kuk-Jin Yoon \orcidlink{0000-0002-1634-2756}}

\authorrunning{Boo et al.}

\institute{KAIST \\
\email{\{boojiyong, byeonginjoung, hyemin0806, kjyoon\}@kaist.ac.kr}}

\maketitle

\section{Overview of Supplementary Material}
This supplementary material provides additional details on the network architecture (\Sref{sec:sup_arch}), training losses (\Sref{sec:sup_loss}), and training details (\Sref{sec:sup_implementation_details}). We further include additional experiments on the Apollo dataset~\cite{guo2020gen} (\Sref{sec:sup_exp_apollo}), an ablation study on the hyperparameter $\theta$ (\Sref{sec:sup_ablation_angle}), a computational efficiency comparison (\Sref{sec:sup_computational_efficiency}), an analysis of the HSDF formulation and the height-directional Eikonal regularization (\Sref{sec:sup_hsdf_eikonal}), an ablation of the LSPE design (\Sref{sec:sup_lspe_ablation}), an evaluation of scalability without dense LiDAR supervision (\Sref{sec:sup_weak_supervision}), and additional qualitative results on the OpenLane~\cite{chen2022persformer} and Apollo datasets (\Sref{sec:sup_additional_qualitative_results}).

\section{Implementation Details}
\label{sec:supp_implementation}

\subsection{Architecture}
\label{sec:sup_arch}

\noindent \textbf{Dense 3D Sampling.} 
In our framework, the number of vertical samples $N(x_i, y_j)$ varies across the BEV grid because the valid vertical search range is defined according to the camera view frustum. Specifically, for each BEV location $(x_i, y_j)$, the vertical interval is clipped by the local bounds $[z_{\min}(x_i), z_{\max}(x_i)]$, which expand with longitudinal distance and therefore produce different valid sampling lengths at different grid cells. As a result, the required number of samples $N(x_i, y_j)$ is determined adaptively for each location. However, directly using a different number of samples for each grid cell leads to irregular tensor shapes. To simplify the implementation, we construct a dense 3D coordinate volume of size $(N_{\max}, H, W)$ using a global maximum number of samples, where $N_{\max}$ corresponds to the largest sampling count across the grid. This allows all BEV locations to share a consistent tensor structure while the effective number of valid samples is still controlled by the adaptive bounds. To achieve this, we introduce a binary padding mask that marks invalid sample locations. In particular, vertical samples $z_k$ that exceed the local upper bound $z_{\max}(x_i)$ are masked out during subsequent rendering operations. In addition, to guarantee at least two valid samples ($N \ge 2$) and avoid degenerate zero-entropy distributions in the softmax-based rendering, the first two vertical samples are always preserved and excluded from the padding mask.

\vspace{2mm}
\noindent \textbf{3D-to-2D Feature Projection.} 
We adopt a pre-trained ResNet-50~\cite{he2016deep} to extract the front-view feature map $\mathcal{F}_{FV}$. Specifically, we utilize the output from the \texttt{conv4\_x} block (corresponding to \texttt{layer3} in PyTorch implementations), which provides a feature map with $C = 1024$ channels and a spatial resolution of $1/16$ of the original input image.  To construct a dense 3D feature volume from the extracted feature map $\mathcal{F}_{FV}$, the sampled 3D points $\mathcal{P}_{sampled}$ are first projected onto the original image plane (resized to $600 \times 800$) using the camera intrinsics and extrinsics. The projected image coordinates are then downscaled by a factor of 16 to align with the spatial dimensions of $\mathcal{F}_{FV}$. Using PyTorch's \texttt{grid\_sample} function with bilinear interpolation, we retrieve the pixel-aligned features for all valid 3D points, yielding a batched 4D feature volume $\mathcal{V}$ of shape $(B, C, N_{\max}, H, W)$, where $B$ is the batch size.

\vspace{2mm}
\noindent \textbf{Implicit HSDF Network.} 
To efficiently predict the HSDF values $\hat{s}$ across the dense 3D feature volume $\mathcal{V}$, we first compress the feature sizes and design a MLP equipped with explicit positional encoding. To mitigate computational costs, we first compress the features from 1024 to 256 channels via a \texttt{Linear} layer, followed by \texttt{LayerNorm} and \texttt{ReLU}. 

To effectively capture high-frequency geometric details of the road surface, we apply the Fourier positional encoding $\gamma(\cdot)$ introduced in NeRF~\cite{mildenhall2021nerf}.

\vspace{2mm}
\noindent \textbf{Lane-aware Semantic Positional Encoding (LSPE).} 
To construct the LSPE, the learnable vector $\mathbf{v} \in \mathbb{R}^{1024}$ is initialized from a normal distribution $\mathcal{N}(0, 0.02^2)$. The derived semantic embedding $\mathbf{p}_{\text{sem}}$ is element-wise added to the geometric embedding $\mathbf{p}_{\text{geo}}$ to form the final positional representation.

\subsection{Training Losses}
\label{sec:sup_loss}
We adopt BEV lane detection losses following HeightLane~\cite{park2025heightlane} with novel objectives for our HSDF and LSPE.

\vspace{2mm}
\noindent \textbf{Segmentation Loss $\mathcal{L}_{\text{seg}}$.} 
The segmentation loss is used to supervise the binary classification of lane existence at each BEV grid cell. Essentially, it trains the network to scan the top-down BEV space and identify exactly which grid cells are occupied by any lane markings. To effectively handle the inherent class imbalance between sparse lane pixels and the background, we combine Binary Cross-Entropy (BCE) and Intersection over Union (IoU) loss:
\begin{align}
\mathcal{L}_{\text{BCE}} &= -\frac{1}{N_{BEV}} \sum \Big[ \mathbf{Y}_{\text{seg}} \log \hat{\mathbf{Y}}_{\text{seg}} + (1-\mathbf{Y}_{\text{seg}}) \log(1-\hat{\mathbf{Y}}_{\text{seg}}) \Big], \\
\mathcal{L}_{\text{IoU}} &= 1 - \frac{\sum \hat{\mathbf{Y}}_{\text{seg}} \mathbf{Y}_{\text{seg}}}{\sum (\hat{\mathbf{Y}}_{\text{seg}} + \mathbf{Y}_{\text{seg}} - \hat{\mathbf{Y}}_{\text{seg}} \mathbf{Y}_{\text{seg}})}, \\
\mathcal{L}_{\text{seg}} &= \mathcal{L}_{\text{BCE}} + \mathcal{L}_{\text{IoU}},
\label{eq:loss_seg}
\end{align}
where $N_{BEV}$ is the total number of grid cells, $\mathbf{Y}_{\text{seg}}$ is the binary ground truth, and $\hat{\mathbf{Y}}_{\text{seg}} = \sigma(\mathbf{P}_{\text{seg}})$ is the predicted probability after the sigmoid activation.

\vspace{2mm}
\noindent \textbf{Offset Loss $\mathcal{L}_{\text{off}}$.} 
Due to the discretization of the BEV grid, predicting exact lane coordinates requires sub-grid refinement. The offset loss aims to recover the precise lateral sub-pixel position (y-offset) of the lane points within each occupied grid cell. Rather than formulating this as an unbounded regression problem, the relative lateral offset from the left edge to the right edge of a cell is normalized to a continuous range of $[0, 1]$. We then treat this normalized offset prediction as a classification-like task using BCE, which naturally restricts the network's output to the cell boundaries. It is computed only for valid lane regions:
\begin{equation}
\mathcal{L}_{\text{off}} = -\frac{1}{\sum \mathbf{Y}_{\text{seg}}} \sum \mathbf{Y}_{\text{seg}} \Big[ \mathbf{Y}_{\text{off}} \log \hat{\mathbf{Y}}_{\text{off}} + (1-\mathbf{Y}_{\text{off}}) \log(1-\hat{\mathbf{Y}}_{\text{off}}) \Big],
\label{eq:loss_off}
\end{equation}
where $\hat{\mathbf{Y}}_{\text{off}}$ is the predicted sub-grid offset probability.

\vspace{2mm}
\noindent \textbf{Embedding Loss $\mathcal{L}_{\text{emb}}$.} 
To differentiate multiple lanes within the same BEV space, we formulate lane clustering as an instance embedding task. We employ a push-pull loss~\cite{neven2018towards} that pulls pixel embeddings $\mathbf{e}_i$ belonging to the same lane instance close to their cluster center $\boldsymbol{\mu}_c$, while pushing the centers of different lane instances apart:
\begin{align}
\mathcal{L}_{\text{pull}}(c) &= \frac{1}{N_c} \sum_{i=1}^{N_c} \max\big(0, \|\mathbf{e}_i - \boldsymbol{\mu}_c\| - \delta_v\big)^2, \\
\mathcal{L}_{\text{push}}(c_A, c_B) &= \max\big(0, \delta_d - \|\boldsymbol{\mu}_{c_A} - \boldsymbol{\mu}_{c_B}\|\big)^2, \\
\mathcal{L}_{\text{emb}} &= \frac{1}{N_{lane}} \sum_{c=1}^{N_{lane}} \mathcal{L}_{\text{pull}}(c) + \frac{1}{N_{lane}(N_{lane}-1)} \sum_{c_A \neq c_B} \mathcal{L}_{\text{push}}(c_A, c_B),
\label{eq:loss_emb}
\end{align}
where $N_{lane}$ is the number of lane instances and $N_c$ is the number of pixels in instance $c$. The variance margin $\delta_v$ and distance margin $\delta_d$ control the clustering tightness and inter-cluster separation, respectively.

\vspace{2mm}
\noindent \textbf{2D Supervision Loss $\mathcal{L}_{\text{2d}}$.} 
To provide direct supervision to the image backbone and stabilize feature learning before the 3D lifting stage, we apply an auxiliary 2D supervision task on the perspective image plane. Specifically, the network predicts a dense 2D lane segmentation map and a corresponding instance embedding map from the front-view feature map. The 2D supervision consists of two components. First, a 2D segmentation loss ($\mathcal{L}_{\text{seg\_2d}}$) identifies lane pixels in the image using a combination of Binary Cross Entropy (BCE) and IoU losses. Second, a 2D instance embedding loss ($\mathcal{L}_{\text{emb\_2d}}$) groups pixels belonging to the same lane using the discriminative push–pull formulation. Therefore, the auxiliary loss is defined as:
\begin{equation}
\mathcal{L}_{\text{2d}} = \lambda_{\text{seg\_2d}}\mathcal{L}_{\text{seg\_2d}} + \lambda_{\text{emb\_2d}}\mathcal{L}_{\text{emb\_2d}}.
\label{eq:loss_2d}
\end{equation}

\vspace{2mm}
\noindent \textbf{Height map Loss $\mathcal{L}_{\mathcal{H}}$.}
The height map loss supervises the road geometry recovered from the HSDF at two levels: the rendered surface and the per-sample distance field. It consists of a rendered height map loss $\mathcal{L}_{\text{render}}$ and an HSDF loss $\mathcal{L}_{\text{hsdf}}$.

The rendered height map loss directly supervises the road height map rendered from the HSDF. We utilize a Smooth L1 loss between the estimated height map $\mathcal{H}$ and the ground truth $\mathcal{H}_{GT}$, computed on valid regions masked by $\mathbf{M}_{\mathcal{H}}$:
\begin{equation}
\mathcal{L}_{\text{render}} = \frac{1}{\sum \mathbf{M}_{\mathcal{H}}} \sum \mathbf{M}_{\mathcal{H}} \cdot \text{SmoothL1}\big(\mathcal{H} - \mathcal{H}_{GT}\big),
\label{eq:loss_render}
\end{equation}
where $\text{SmoothL1}(d) = 0.5\,d^2$ if $|d| < 1$ and $|d| - 0.5$ otherwise.

In addition to the rendered surface, we directly supervise the predicted HSDF value at each sample point. The ground-truth HSDF at sample $(x_i, y_j, z_k)$ is defined as $s_{GT} = z_k - \mathcal{H}_{GT}(x_i, y_j)$. We apply a Smooth L1 loss between the predicted value $\hat{s}$ and $s_{GT}$ over valid samples masked by $\mathbf{M}_s$, which combines the valid ground-truth region with the per-sample validity mask:
\begin{equation}
\mathcal{L}_{\text{hsdf}} = \frac{1}{\sum \mathbf{M}_s} \sum \mathbf{M}_s \cdot \text{SmoothL1}\big(\hat{s} - s_{GT}\big).
\label{eq:loss_hsdf}
\end{equation}
This point-wise supervision enforces a geometrically consistent distance field across the entire sampled volume, complementing the surface-level supervision of $\mathcal{L}_{\text{render}}$.

The total height map loss is the sum of the two terms:
\begin{equation}
\mathcal{L}_{\mathcal{H}} = \mathcal{L}_{\text{render}} + \mathcal{L}_{\text{hsdf}}.
\label{eq:loss_height}
\end{equation}

\vspace{2mm}
\noindent \textbf{Height-directional Eikonal Regularization $\mathcal{L}_{\text{eik}}$.} 
To ensure the HSDF estimator learns a reliable distance field, we introduce the novel Height-directional Eikonal constraint. As our HSDF represents distances along the vertical direction, the gradient magnitude along the $z$-axis is theoretically expected to be $1$. Approximating this property using finite differences across adjacent valid samples, we formulate the loss as:
\begin{equation}
\mathcal{L}_{\text{eik}} = \frac{1}{\sum \mathbf{M}_\mathcal{H}} \sum_{(i,j,k)} \mathbf{M}_\mathcal{H} \left| \frac{\hat{s}(x_i,y_j,z_{k+1})-\hat{s}(x_i,y_j,z_k)} {z_{k+1}-z_k} - 1 \right|.
\label{eq:eikonal}
\end{equation}

\vspace{2mm}
\noindent \textbf{Heatmap Loss $\mathcal{L}_{\text{hm}}$.} 
The heatmap loss optimizes the lane heatmap $\mathcal{M}$ predicted from the HSDF features $\mathcal{F}_{hsdf}$. Since the BEV space is heavily dominated by non-lane pixels, we employ a Gaussian focal loss~\cite{duan2019centernet} to provide soft supervision and mitigate the resulting class imbalance:
\begin{equation}
\mathcal{L}_{\text{hm}} = -\frac{1}{N_{\text{pos}}} \sum \begin{cases}
(1 - \mathcal{M})^\alpha \log(\mathcal{M}), & \text{if } \mathcal{M}_{GT} = 1 \\
(1 - \mathcal{M}_{GT})^\beta \mathcal{M}^\alpha \log(1 - \mathcal{M}), & \text{otherwise}
\end{cases}
\label{eq:loss_hm}
\end{equation}
where $N_{\text{pos}}$ is the number of positive lane center points, and the hyper-parameters $\alpha$ and $\beta$ control the penalties for easy and hard examples.

\subsection{Training Details}
\label{sec:sup_implementation_details}
Our HSDF-Lane model is trained end-to-end using the AdamW optimizer with a weight decay of $10^{-2}$ and a cosine annealing learning rate scheduler. The initial learning rate is set to $5 \times 10^{-4}$, preceded by a linear warmup over the first $1000$ iterations that increases the rate from zero to this value. To ensure full convergence across different dataset scales, the training process spans $24$ epochs for the OpenLane dataset~\cite{chen2022persformer} and $100$ epochs for the Apollo 3D synthetic dataset~\cite{guo2020gen}. We use a total batch size of $32$ distributed across two NVIDIA H200 GPUs. The weighting hyperparameters are empirically set as follows: $\lambda_{\text{seg}} = 5$, $\lambda_{\text{off}} = 60$, $\lambda_{\text{emb}} = 1$, $\lambda_{\text{2d}} = 1$\ ($\lambda_{\text{seg\_2d}} = 3$, $\lambda_{\text{emb\_2d}} = 0.5$), $\lambda_{\text{geo}} = 10$, $\lambda_{\text{eik}} = 0.1$ and $\lambda_{\text{hm}} = 5$.

\section{Additional Experiments}
\label{sec:supp_additional_exp}
In this section, we provide additional experimental analyses to comprehensively evaluate the proposed HSDF-Lane. Specifically, we present: 1) experimental results on the Apollo 3D Synthetic dataset~\cite{guo2020gen}, 2) an ablation study analyzing the impact of the spatial sampling hyperparameter, 3) an analysis of computational efficiency, and 4) additional qualitative visualizations in complex driving scenarios.

\subsection{Results on Apollo Dataset}
\label{sec:sup_exp_apollo}

\noindent \textbf{Dataset and Setup.} 
The Apollo 3D Synthetic dataset~\cite{guo2020gen} is a synthetic dataset built using the Unity game engine to simulate diverse 3D autonomous driving worlds. It comprises a total of 10,500 images rendered from highway, urban, and residential maps based on real regions in Silicon Valley, including varying day-times, camera heights, and pitch angles. Based on specific evaluation criteria, the dataset provides three distinct splits: \textit{Balanced Scenes} for standard benchmarking, \textit{Rarely Observed Scenes} to test generalization to unfamiliar urban layouts and sharp turnings, and \textit{Scenes with Visual Variations} to evaluate robustness against illumination changes.

Because the Apollo dataset does not provide LiDAR point clouds, we construct the ground-truth road height maps by leveraging the provided depth maps and 2D semantic segmentation data. For certain scenes where explicit road segmentation annotations are missing, we utilize the Segment Anything Model (SAM)~\cite{kirillov2023segany} to automatically generate pseudo road masks. Specifically, we project the 3D lane annotations onto the 2D image plane and sample points along these projected lanes to serve as positive prompts for SAM. To spatially constrain the road surface extraction, we utilize points from existing object masks and the top 25\% region of the image as negative prompts, effectively suppressing background regions.

\begin{table}[t]
  \caption{\textbf{Quantitive results on Apollo 3D Synthetic dataset~\cite{guo2020gen}.} The best results are highlighted in \textbf{bold}, while the second-best results are \underline{underlined}.
  }
  \label{tab:supple_02_apollo}
  \centering
  \resizebox{1.0\columnwidth}{!}{%
  \begin{tabular}{c|l|c|c|cc|cc}
    \hline
    \multirow{2}{*}{Scene} & \multirow{2}{*}{Method} & \multirow{2}{*}{F1-Score(\%)$\uparrow$} & \multirow{2}{*}{AP(\%)$\uparrow$} & \multicolumn{2}{c|}{X-error} & \multicolumn{2}{c}{Z-error} \\
    \cline{5-8}
     & & & & near (m)$\downarrow$ & far (m)$\downarrow$ & near (m)$\downarrow$ & far (m)$\downarrow$ \\
    \hline
     \multirow{8}{*}{\rotatebox{90}{\textit{Balanced Scene}}} & PersFormer~\cite{chen2022persformer} & 92.9 & - & 0.054 & 0.356 & \underline{0.010} & 0.234 \\
     & BEVLaneDet~\cite{wang2023bev} & 96.9 & - & \textbf{0.016} & 0.242 & 0.020 & 0.216 \\
     & LaneCPP~\cite{pittner2024lanecpp}& 97.4 & \textbf{99.5} & 0.030 & 0.277 & 0.011 & 0.206 \\
     & LATR~\cite{luo2023latr}& 96.8 & 97.9 & 0.022 & 0.253 &\textbf{ 0.007} & 0.202 \\
     & DV-3DLane~\cite{luo2024dv}& 96.4 & 97.6 & 0.046 & 0.299 & 0.016 & 0.213 \\
     & GLane3D-B~\cite{ozturk2025glane3d}& \underline{98.1} & 98.8 & \underline{0.021} & 0.250 & \textbf{0.007} & 0.213\\
     & HeightLane~\cite{park2025heightlane}&  97.9 & \underline{98.9} & 0.022 & \underline{0.075} & 0.051 & \underline{0.133}  \\
     & \cellcolor{gray!30}\textbf{HSDF-Lane (Ours)}& \cellcolor{gray!30}\textbf{98.8} &
     \cellcolor{gray!30}\textbf{99.5}&
     \cellcolor{gray!30}\underline{0.021} & \cellcolor{gray!30}\textbf{0.072} & \cellcolor{gray!30}0.041 & \cellcolor{gray!30}\textbf{0.095}  \\
    \hline
    \multirow{8}{*}{\rotatebox{90}{\textit{Rare Scene}}} & PersFormer~\cite{chen2022persformer} & 87.5 & - & 0.107 & 0.782 & 0.024 & 0.602 \\
     & BEVLaneDet~\cite{wang2023bev} & 97.6 & - & \textbf{0.031} & 0.594 & 0.040 & 0.556 \\
     & LaneCPP~\cite{pittner2024lanecpp}& 96.2 & 98.6 & 0.073 & 0.651 & \underline{0.023} & 0.543 \\
     & LATR~\cite{luo2023latr}& 96.1 & 97.3 & 0.050 & 0.600 & \textbf{0.015} & 0.532 \\
     & DV-3DLane~\cite{luo2024dv}& 95.5 & 97.2 & 0.071 & 0.664 & 0.025 & 0.568 \\
     & GLane3D-B~\cite{ozturk2025glane3d}& \underline{98.4} & \underline{99.1} & 0.044 & 0.621 & \underline{0.023} & 0.566 \\
     & HeightLane~\cite{park2025heightlane}& 95.3 & 97.5 & 0.037 & \underline{0.064} & 0.104 & \underline{0.272} \\
     & \cellcolor{gray!30}\textbf{HSDF-Lane (Ours)}& \cellcolor{gray!30}\textbf{99.2} & \cellcolor{gray!30}\textbf{99.6} & \cellcolor{gray!30}\underline{0.034} & \cellcolor{gray!30}\textbf{0.061} & \cellcolor{gray!30}0.065 & \cellcolor{gray!30}\textbf{0.134} \\
    \hline
    \multirow{8}{*}{\rotatebox{90}{\textit{Visual Variations}}} & PersFormer~\cite{chen2022persformer} & 89.6 & - & 0.074 & 0.430 & \textbf{0.015} & 0.266 \\
     & BEVLaneDet~\cite{wang2023bev} & 95.0 & - & \underline{0.027} & 0.320 & 0.031 & 0.256 \\
     & LaneCPP~\cite{pittner2024lanecpp}& 90.4 & 93.7 & 0.054 & 0.327 & 0.020 & 0.222 \\
     & LATR~\cite{luo2023latr}& 95.1 & 96.6 & 0.045 & 0.315 & \underline{0.016} & 0.228 \\
     & DV-3DLane~\cite{luo2024dv}& 91.3 & 93.4 & 0.095 & 0.417 & 0.040 & 0.320\\
     & GLane3D-B~\cite{ozturk2025glane3d}& 92.7 & 94.8 & 0.046 & 0.364 & 0.020 & 0.317 \\
     & HeightLane~\cite{park2025heightlane}& \underline{97.0} & \underline{98.6} &\textbf{ 0.026} & \underline{0.109} & 0.055 & \underline{0.137}\\
     & \cellcolor{gray!30}\textbf{HSDF-Lane (Ours)}& \cellcolor{gray!30}\textbf{97.9} & \cellcolor{gray!30}\textbf{99.0} & \cellcolor{gray!30}\textbf{0.026} & \cellcolor{gray!30}\textbf{0.103} & \cellcolor{gray!30}0.048 & \cellcolor{gray!30}\textbf{0.113} \\
    \hline
  \end{tabular}%
  }
\end{table}

\begin{table}[ht]
\centering
\caption{\textbf{Comparison of height map estimation on Apollo 3D Synthetic dataset~\cite{guo2020gen}.} The best results for each scenario are highlighted in \textbf{bold}.}
\label{tab:apollo_height}

\newcolumntype{C}{>{\centering\arraybackslash}p{1.2cm}} 

\resizebox{0.9\linewidth}{!}{%
\begin{tabular}{c|c|CC|CCC}
\hline
\multirow{2}{*}{\textbf{Scene}} & \multirow{2}{*}{\textbf{Method}} & \multicolumn{2}{c|}{\textbf{Error Metrics} ($\downarrow$)} & \multicolumn{3}{c}{\textbf{Accuracy Metrics} ($\uparrow$)} \\ \cline{3-7} 
 & & \textbf{MAE} & \textbf{RMSE} & \textbf{@0.05} & \textbf{@0.1} & \textbf{@0.2} \\ \hline
 
 & HeightLane~\cite{park2025heightlane} & 0.063 & 0.157 & 0.762 & 0.859 & 0.923 \\
\rowcolor{gray!20} \cellcolor{white}\multirow{-2}{*}{\textit{Balanced Scene}} & HSDF-Lane & \textbf{0.026} & \textbf{0.056} & \textbf{0.849} & \textbf{0.946} & \textbf{0.991} \\ \hline

 & HeightLane~\cite{park2025heightlane} & 0.098 & 0.188 & 0.573 & 0.736 & 0.863 \\
\rowcolor{gray!20} \cellcolor{white}\multirow{-2}{*}{\textit{Rare Scene}} & HSDF-Lane & \textbf{0.048} & \textbf{0.095} & \textbf{0.707} & \textbf{0.875} & \textbf{0.966} \\ \hline

 & HeightLane~\cite{park2025heightlane} & 0.054 & 0.145 & 0.777 & 0.881 & 0.943 \\
\rowcolor{gray!20} \cellcolor{white}\multirow{-2}{*}{\textit{Visual Variations}} & HSDF-Lane & \textbf{0.032} & \textbf{0.082} & \textbf{0.835} & \textbf{0.932} & \textbf{0.980} \\ \hline
\end{tabular}%
}
\end{table}

\noindent \textbf{Quantitative Results.} 
We evaluate the 3D lane detection performance using F1-Score, X-error, and Z-error, consistent with the OpenLane benchmark~\cite{chen2022persformer}, alongside Average Precision (AP), which is specifically employed for the Apollo dataset. As shown in \Tref{tab:supple_02_apollo}, HSDF-Lane consistently achieves state-of-the-art performance across all three dataset splits. In the \textit{Balanced Scene}, our method outperforms the baseline HeightLane~\cite{park2025heightlane}. The performance gap significantly widens in more challenging scenarios. In the \textit{Rare Scene} and \textit{Visual Variations} splits, our approach demonstrates superior generalization capability to unfamiliar road topologies and varying lighting conditions. Notably, our method drastically reduces the far-range Z-error in the \textit{Rare Scene}.

Furthermore, \Tref{tab:apollo_height} reports the height map estimation metrics. HSDF-Lane consistently reduces both MAE and RMSE across all evaluation splits. Notably, HSDF-Lane shows clear improvements in the \textit{Rare Scene} subset. These results demonstrate that the proposed HSDF modeling produces a more accurate and geometrically consistent road surface compared to prior anchor-based direct regression approaches.

\begin{figure}[h]
  \centering
  \includegraphics[width=0.8\textwidth]{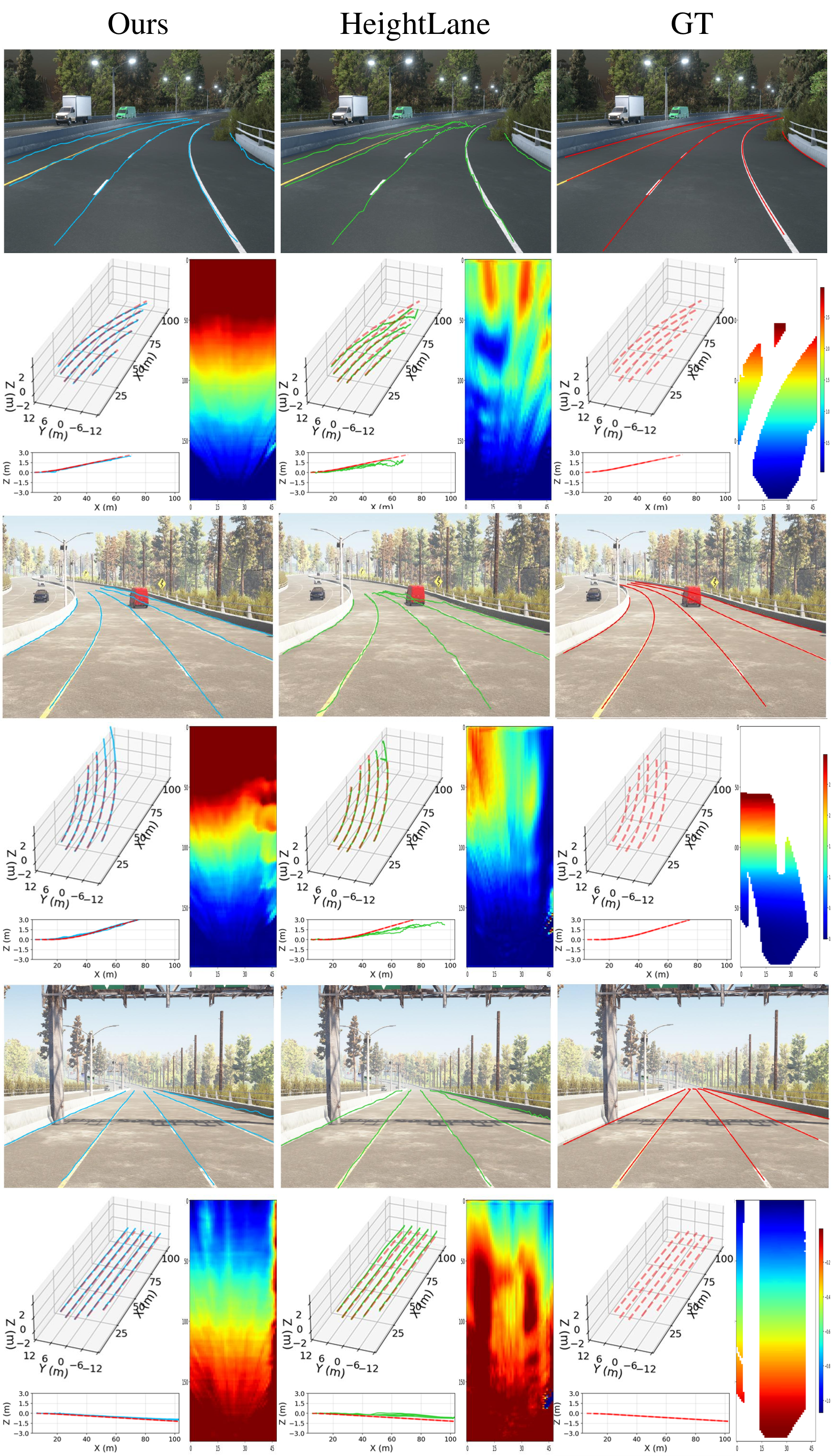} 
  \caption{\textbf{Qualitative results on Apollo 3D Synthetic dataset~\cite{guo2020gen}.} We compare our proposed HSDF-Lane against HeightLane~\cite{park2025heightlane}. The visualizations illustrate the 3D lane projections onto the input image, the reconstructed 3D lanes, and the corresponding estimated height maps. The Ground Truth (GT) is displayed on the far left, with GT lanes denoted by red dashed lines.}
  \label{fig:sp_ap_1}
\end{figure}

\begin{figure}[h]
  \centering
  \includegraphics[width=0.8\textwidth]{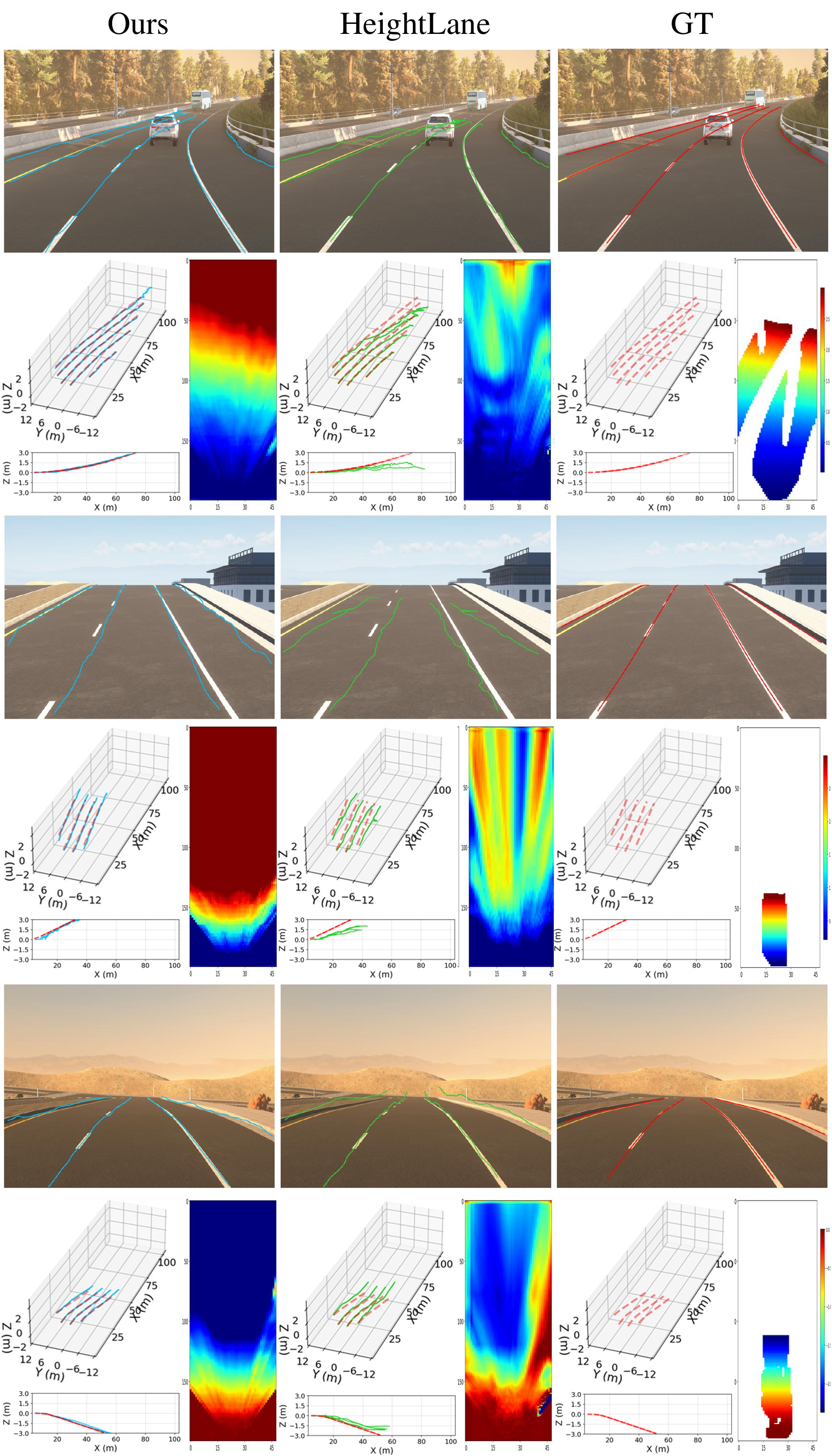} 
  \caption{\textbf{Qualitative results on Apollo 3D Synthetic dataset~\cite{guo2020gen}.} We compare our proposed HSDF-Lane against HeightLane~\cite{park2025heightlane}. The visualizations illustrate the 3D lane projections onto the input image, the reconstructed 3D lanes, and the corresponding estimated height maps. The Ground Truth (GT) is displayed on the far left, with GT lanes denoted by red dashed lines.}
  \label{fig:sp_ap_2}
\end{figure}

\vspace{2mm}
\noindent \textbf{Qualitative Results.}
The qualitative results on the Apollo dataset~\cite{guo2020gen} are shown in \Fref{fig:sp_ap_1} and \Fref{fig:sp_ap_2}.
 HSDF-Lane robustly reconstructs 3D lanes under varying gradients and illumination conditions. Unlike prior methods that suffer from height fluctuations, our approach maintains geometric consistency to yield smooth predictions, demonstrating the effectiveness of the proposed HSDF and LSPE.

\subsection{Effect of the Maximum Slope Angle $\theta$}
\label{sec:sup_ablation_angle}

\noindent \textbf{Ablation Setup.} 
In our dense 3D sampling strategy, the maximum slope angle $\theta$ is a crucial hyperparameter. A smaller $\theta$ restricts the search space, potentially truncating valid road surfaces on steep inclines, whereas a larger $\theta$ degrades spatial resolution and introduces noise unless sampling density is proportionally increased. To evaluate the impact of $\theta$, we performed an ablation study on the OpenLane-300 subsets~\cite{chen2022persformer} by varying $\theta \in \{4^\circ, 5^\circ, 6^\circ\}$ and comparing our proposed HSDF-Lane with HeightLane~\cite{park2025heightlane}.

\noindent \textbf{Results.} 
As summarized in \Tref{tab:supple_00_abl}, HSDF-Lane demonstrates strong robustness to different vertical sampling bounds and consistently achieves better height estimation accuracy than the baseline. The best overall performance is obtained when $\theta = 5^\circ$, which is the configuration adopted in the main manuscript.

\begin{table}[h]
  \caption{\textbf{Ablation study of the parameter $\theta$ on the OpenLane 300 subset~\cite{chen2022persformer}.} The best results are highlighted in \textbf{bold}.
  }
  \label{tab:supple_00_abl}
  \centering
  \resizebox{0.9\columnwidth}{!}{%
  \begin{tabular}{c|c|c|cc|cc|cc|ccc}
    \hline
    \multirow{2}{*}{\textbf{Method}} & \multirow{2}{*}{$\theta$} & \multirow{2}{*}{F1-Score(\%)$\uparrow$} & \multicolumn{2}{c|}{X-error (m)$\downarrow$} & \multicolumn{2}{c|}{Z-error (m)$\downarrow$}  & \multicolumn{2}{c|}{$\mathcal{H}$ Error $\downarrow$}  & \multicolumn{3}{c}{ $\mathcal{H}$ Accuracy $\uparrow$}\\
    \cline{4-12}
     & & & near & far & near & far & MAE & RMSE & @0.05 & @0.1 & @0.2 \\
    \hline
    HeightLane &  & 74.8 & 0.219 & 0.289 & 0.205 & 0.238 & 0.178 & 0.327 & 0.265 & 0.474 & 0.730 \\
    \rowcolor{gray!20}HSDF-Lane & \cellcolor{white}\multirow{-2}{*}{4} & \textbf{75.6} & 0.229 & 0.286 & 0.198 & 0.149 & 0.169 & 0.332 & 0.260 & 0.472 & 0.745 \\
    \hline
    HeightLane & & 73.3 & 0.225 & 0.293 & 0.204 & 0.236 & 0.181 & 0.351 & 0.272 & 0.482 & 0.736\\
    \rowcolor{gray!20}HSDF-Lane & \cellcolor{white}\multirow{-2}{*}{5} & \textbf{75.6} & \textbf{0.208} & \textbf{0.268} & \textbf{0.159} & \textbf{0.146} & 0.158 & 0.326 & \textbf{0.288} & \textbf{0.503} & \textbf{0.767}\\
    \hline
    HeightLane & & 75.3 & 0.239 & 0.284 & 0.213 & 0.246 & 0.184 & 0.336 & 0.262 & 0.472 & 0.727\\
    \rowcolor{gray!20}HSDF-Lane & \cellcolor{white}\multirow{-2}{*}{6} & 74.8 & 0.231 & 0.283 & 0.174 & 0.161 & \textbf{0.157} & \textbf{0.284} & \textbf{0.288} & 0.498 & 0.749\\
    \hline
  \end{tabular}%
  }
\end{table}

\subsection{Computational Efficiency}
\label{sec:sup_computational_efficiency}

\begin{table}[h]
\centering
\caption{\textbf{Efficiency comparison.} We evaluate the F1-score, inference speed (FPS), and computational overhead (FLOPs and parameter count) for each method, measured on an RTX 3090 with a batch size of 1. The best results are highlighted in \textbf{bold}. $({\S})$: Reproduced due to lack of official code. (${\dagger}$): HSDF-Lane equipped with FPN.}

\label{tab:supple_eff}
\begin{tabular}{l|c|c|c|c|c}
    \hline

    \textbf{Methods} & \textbf{Backbone} & \textbf{F1-Score} $\uparrow$ & \textbf{FPS} $\uparrow$ & \textbf{FLOPs} & \textbf{\# of Params.} \\ \hline
    LATR~\cite{luo2023latr} & ResNet-50 & 61.9 & 15.92 & 135.87G & 46.84M \\
    HeightLane~\cite{park2025heightlane} & ResNet-50 & 62.7 &  17.24 & 417.58G & 73.40M \\
    SC-Lane~\cite{park2025sc} & ResNet-50 & 64.3 &  18.02 & 375.10G & 70.10M \\
    SparseLaneSTP$^{\S}$~\cite{pittner2025sparselanestp} & ResNet-50 & 66.1 &  11.02 & 147.46G & 51.54M \\
    \rowcolor{gray!20} \textbf{HSDF-Lane} & ResNet-50 & \underline{66.3} & \underline{18.76} & 269.55G & 28.35M\\
    \rowcolor{gray!20} \textbf{HSDF-Lane}$^{\dagger}$ & ResNet-50 & \textbf{66.9} & \textbf{23.19} & 145.08G & 31.25M\\
    \hline
\end{tabular}
\end{table}
\noindent \textbf{Evaluation Setup.} 
To evaluate the practicality of HSDF-Lane for real-world autonomous driving, we analyze computational efficiency in terms of inference speed (FPS), floating-point operations (FLOPs), and total parameter count. The comparison includes representative state-of-the-art 3D lane detection models~\cite{luo2023latr, park2025heightlane, park2025sc, pittner2025sparselanestp}. All inference speeds are measured on a single NVIDIA RTX 3090 GPU with a batch size of 1. To ensure stable timing measurements, we first run 100 warm-up iterations, followed by 500 iterations to compute the average FPS.

\noindent \textbf{Results.} 
As summarized in \Tref{tab:supple_eff}, the proposed framework operates at the highest inference speed while maintaining superior performance among the compared methods.


\subsection{Effectiveness of the HSDF Formulation and Eikonal Regularization}
\label{sec:sup_hsdf_eikonal}
Unlike conventional SDFs that model an isotropic field in full 3D space, HSDF adopts a 2.5D formulation that captures continuous road geometry on a BEV grid while substantially reducing 3D complexity. By constraining the Eikonal loss to the vertical derivative ($\partial s/\partial z = 1$), the surface is regularized along the height direction with minimal computational overhead.

To verify that HSDF is more than a reparameterization of direct height regression, we compare the two formulations on the full OpenLane dataset~\cite{chen2022persformer} in \Tref{tab:rebuttal_eikonal}. The HSDF-based formulation yields consistent gains across all error and accuracy metrics over direct regression. Adding the height-directional Eikonal regularization $\mathcal{L}_{\text{eik}}$ further improves geometric consistency, with the most notable gains in the far range and the accuracy-based metrics.

\begin{table}[h]
\centering
\caption{Ablation on the HSDF formulation and the height-directional Eikonal regularization $\mathcal{L}_{\text{eik}}$, evaluated on height map $\mathcal{H}$ prediction over the full OpenLane dataset~\cite{chen2022persformer}. All variants are trained with height map estimation supervision only.}
\label{tab:rebuttal_eikonal}
\resizebox{0.8\columnwidth}{!}{%
    \begin{tabular}{c|cc|cc|cc|ccc}
    \hline
    \multirow{2}{*}{\textbf{Method}} & \multicolumn{2}{c|}{\textbf{Overall} ($\downarrow$)} & \multicolumn{2}{c|}{\textbf{Close} ($\downarrow$)} & \multicolumn{2}{c|}{\textbf{Far} ($\downarrow$)} & \multicolumn{3}{c}{\textbf{Accuracy Metrics} ($\uparrow$)} \\ \cline{2-10}
     & \textbf{MAE} & \textbf{RMSE} & \textbf{MAE} & \textbf{RMSE} & \textbf{MAE} & \textbf{RMSE} & \textbf{@0.05} & \textbf{@0.1} & \textbf{@0.2} \\ \hline
    Direct regression & 0.183 & 0.524 & 0.144 & 0.244 & 0.231 & 0.737 & 0.252 & 0.457 & 0.730 \\ \hline
    HSDF w/o $\mathcal{L}_{\text{eik}}$ & 0.159 & 0.347 & \textbf{0.123} & \textbf{0.208} & 0.204 & 0.465 & 0.293 & 0.510 & 0.761 \\ \hline
    HSDF (Ours) & \textbf{0.157} & \textbf{0.331} & 0.124 & 0.210 & \textbf{0.199} & \textbf{0.436} & \textbf{0.295} & \textbf{0.511} & \textbf{0.762} \\ \hline
    \end{tabular}
}
\end{table}

\subsection{Ablation of the LSPE Design}
\label{sec:sup_lspe_ablation}
We compare our LSPE design with alternative ways of injecting the lane heatmap into the transformer queries on the full OpenLane dataset~\cite{chen2022persformer}, as reported in \Tref{tab:rebuttal_LSPE}. Encoding the heatmap with a full learnable MLP (B) introduces noise that interferes with the geometric positional embedding $\mathbf{p}_{\text{geo}}$, performing even below the no-LSPE baseline (A). Directly concatenating the heatmap (C) also underperforms, as the single-channel semantic signal is overwhelmed by the high-dimensional $\mathbf{p}_{\text{geo}}$. In contrast, our design (D) modulates a single learnable vector with the heatmap probability, acting as a clean attention gate that injects the semantic prior without introducing conflicting signals.

\begin{table}[t]
\centering
\caption{Ablation of LSPE design choices on the full OpenLane dataset~\cite{chen2022persformer} (F1-Score, \%).}
\label{tab:rebuttal_LSPE}
\resizebox{0.7\columnwidth}{!}{%
  \begin{tabular}{@{}l|ccccccc@{}}
    \toprule
    Method & All & \makecell{Up \& \\ Down} & Curve & \makecell{Extreme \\ Weather} & Night & Intersection & \makecell{Merge \\ \& Split}\\
    \midrule
    A: no LSPE & 65.9 & 56.3 & 72.8 & 56.9 & \textbf{58.6} & 58.0 & 64.7 \\
    B: MLP & 65.7 & 55.4 & 72.8 & 57.7 & 58.2 & 57.7 & 63.0 \\
    C: Concat & 66.0 & 57.8 & 72.9 & 59.6 & 57.4 & 58.2 & 64.1 \\
    D: Ours & \textbf{66.3} & \textbf{58.9} & \textbf{73.7} & \textbf{60.7} & 57.4 & \textbf{58.3} & \textbf{65.7} \\
    \bottomrule
  \end{tabular}%
}
\end{table}

\subsection{Scalability without Dense LiDAR Supervision}
\label{sec:sup_weak_supervision}
The main experiments use dense ground-truth height maps accumulated from LiDAR point clouds. To assess whether HSDF-Lane remains applicable under weaker supervision, we retrain the model on the full OpenLane dataset~\cite{chen2022persformer} using pseudo height maps derived from (E) 3D lane annotations only like~\cite{pittner2024lanecpp} and (F) 3D lane annotations combined with raw single-frame LiDAR. As shown in \Tref{tab:rebuttal_weak_supervision}, both variants achieve F1-scores on par with the dense-supervision setting (G), indicating that the proposed framework does not strictly depend on dense LiDAR-derived height maps. As expected, the dense-supervision setting still attains the best height estimation accuracy, since its supervision is directly aligned with the height metrics.

\begin{table}[t]
\centering
\caption{Comparison of height map supervision sources on the full OpenLane dataset~\cite{chen2022persformer}.}
\label{tab:rebuttal_weak_supervision}
\resizebox{\columnwidth}{!}{%
    \begin{tabular}{c|c|cc|cc|cc|ccc}
    \hline
    \multirow{2}{*}{\textbf{Method}} & \multirow{2}{*}{\textbf{F1(\%)}} & \multicolumn{2}{c|}{\textbf{X error} ($\downarrow$)} & \multicolumn{2}{c|}{\textbf{Z error} ($\downarrow$)} & \multicolumn{2}{c|}{\textbf{$\mathcal{H}$ Error} ($\downarrow$)} & \multicolumn{3}{c}{\textbf{Accuracy Metrics} ($\uparrow$)} \\ \cline{3-11}
     & & near & far & near & far & MAE & RMSE & @0.05 & @0.1 & @0.2 \\ \hline
    E: Lane-derived & 66.0 & 0.198 & 0.220 & 0.090 & \textbf{0.113} & 0.279 & 0.549 & 0.224 & 0.399 & 0.622 \\ \hline
    F: Raw LiDAR + Lane & \textbf{66.7} & \textbf{0.194} & \textbf{0.211} & 0.090 & 0.155 & 0.322 & 0.643 & 0.212 & 0.378 & 0.599 \\ \hline
    G: Dense LiDAR (Ours) & 66.3 & 0.201 & 0.223 & \textbf{0.088} & 0.114 & \textbf{0.158} & \textbf{0.313} & \textbf{0.294} & \textbf{0.512} & \textbf{0.763} \\ \hline
    \end{tabular}
}
\end{table}


\subsection{Additional Qualitative Results.}
\label{sec:sup_additional_qualitative_results}
\begin{figure}[t]
  \centering
  \includegraphics[width=1.0\textwidth]{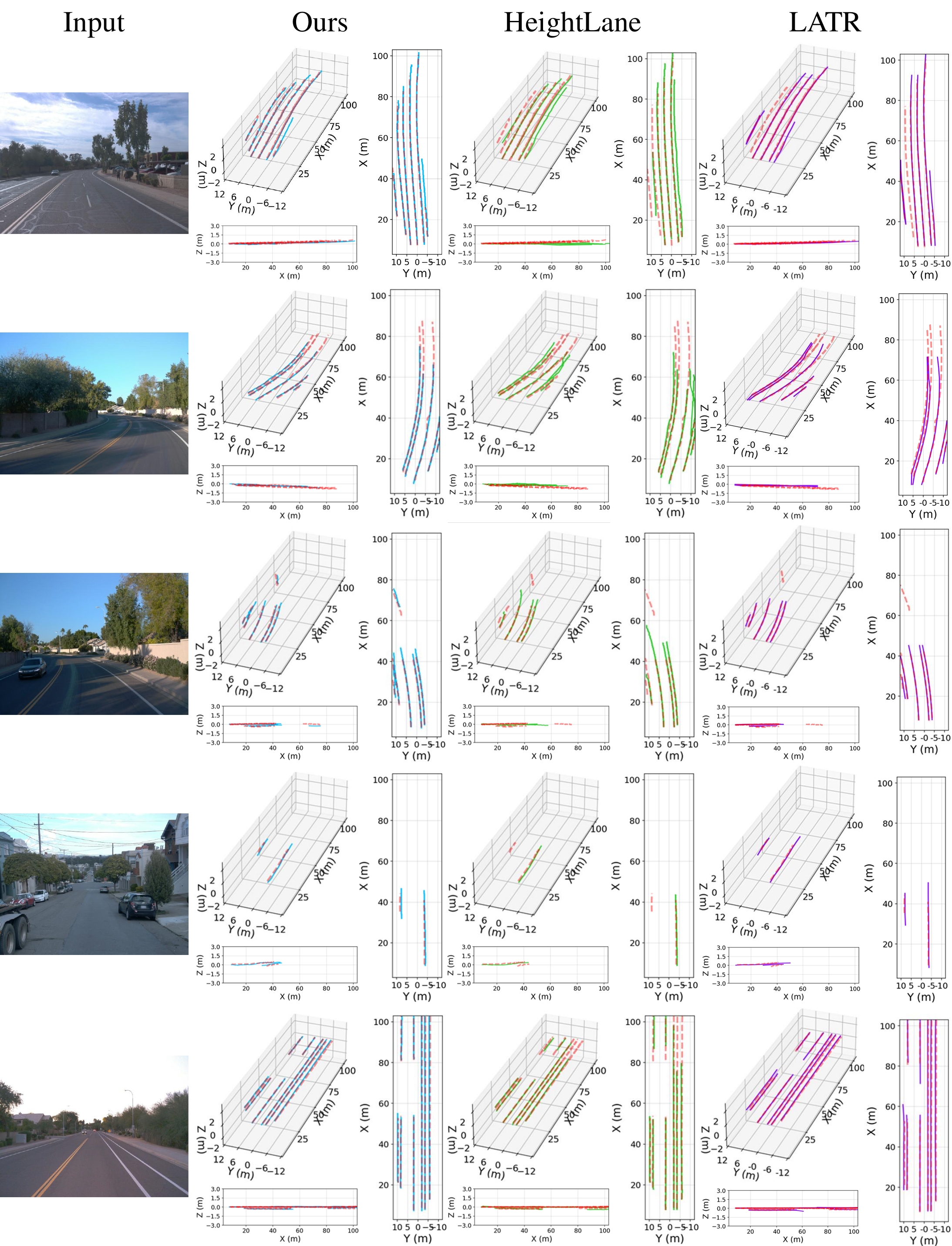} 
  \caption{\textbf{Qualitative result of 3D lane detection on OpenLane Benchmark~\cite{chen2022persformer}.} From left to right: input image $\mathcal{I}$, and 3D lane predictions of \textcolor{blue!60}{HSDF-Lane (ours)}, \textcolor{green}{HeightLane}~\cite{park2025heightlane}, and \textcolor{purple!60}{LATR}~\cite{luo2023latr}. The \textcolor{red}{ground truth (GT)} is indicated by red dashed lines.}
  \label{fig:sp_q_1}
\end{figure}

\begin{figure}[t]
  \centering
  \includegraphics[width=1.0\textwidth]{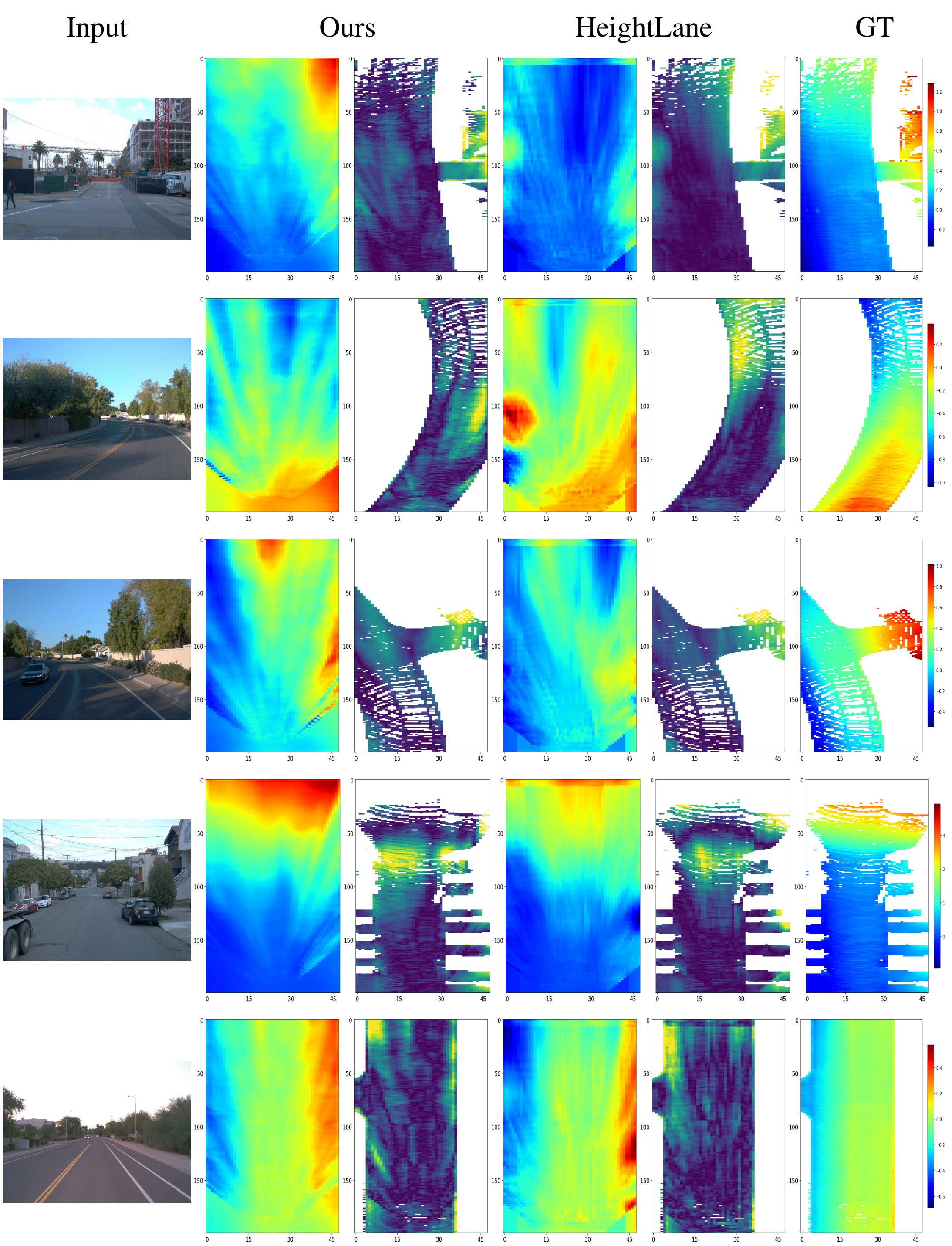} 
  \caption{\textbf{Qualitative comparison of height map estimation on OpenLane Benchmark~\cite{chen2022persformer}.} From left to right: input image $\mathcal{I}$, estimated height maps $\mathcal{H}$ and their corresponding error maps (masked by valid GT regions) for our method and HeightLane~\cite{park2025heightlane}, alongside the GT height map.}
  \label{fig:sp_h_1}
\end{figure}

\begin{figure}[t]
  \centering
  \includegraphics[width=1.0\textwidth]{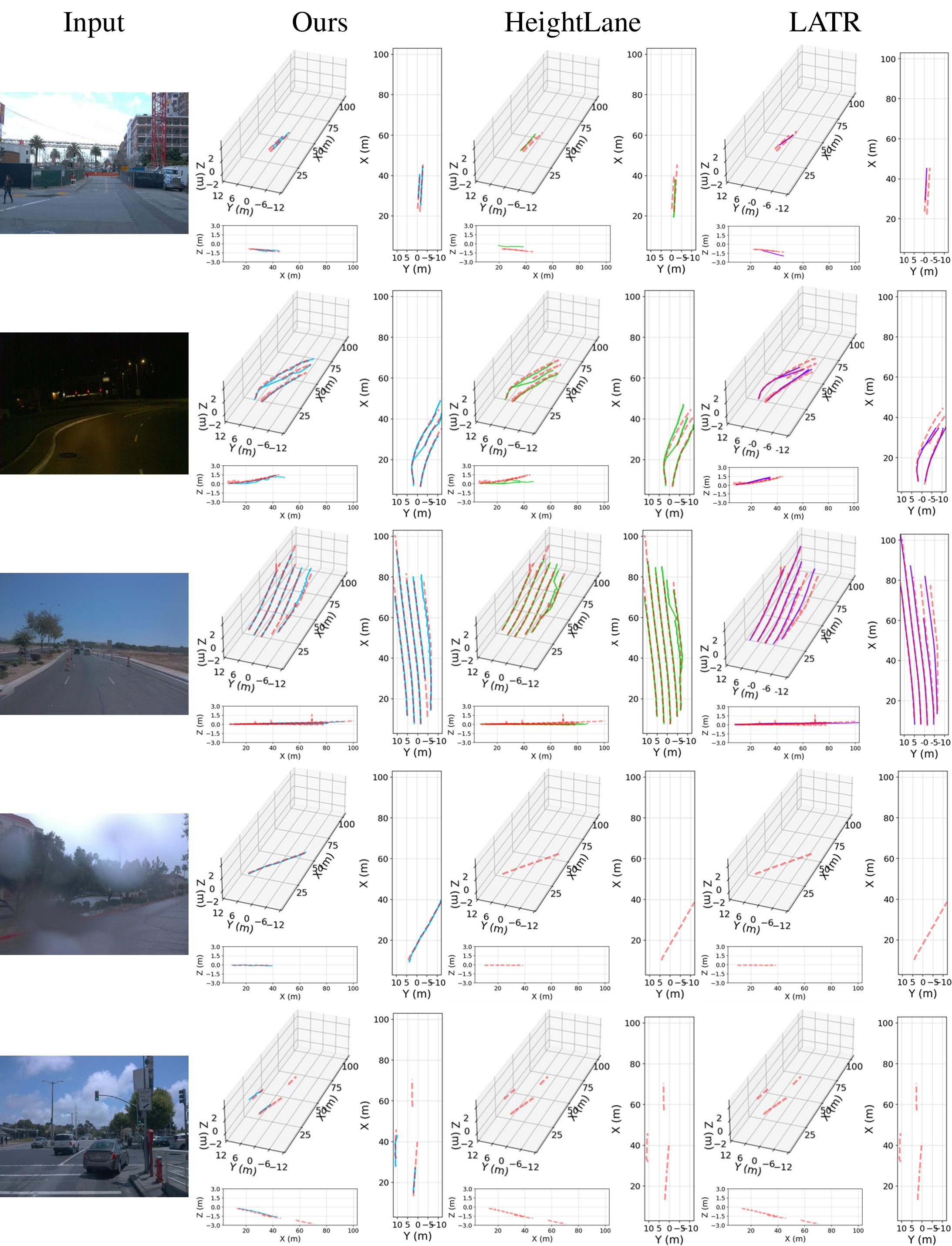} 
  \caption{\textbf{Qualitative result of 3D lane detection on OpenLane Benchmark~\cite{chen2022persformer}.} From left to right: input image $\mathcal{I}$, and 3D lane predictions of \textcolor{blue!60}{HSDF-Lane (ours)}, \textcolor{green}{HeightLane}~\cite{park2025heightlane}, and \textcolor{purple!60}{LATR}~\cite{luo2023latr}. The \textcolor{red}{ground truth (GT)} is indicated by red dashed lines.}
  \label{fig:sp_q_2}
\end{figure}

\begin{figure}[t]
  \centering
  \includegraphics[width=1.0\textwidth]{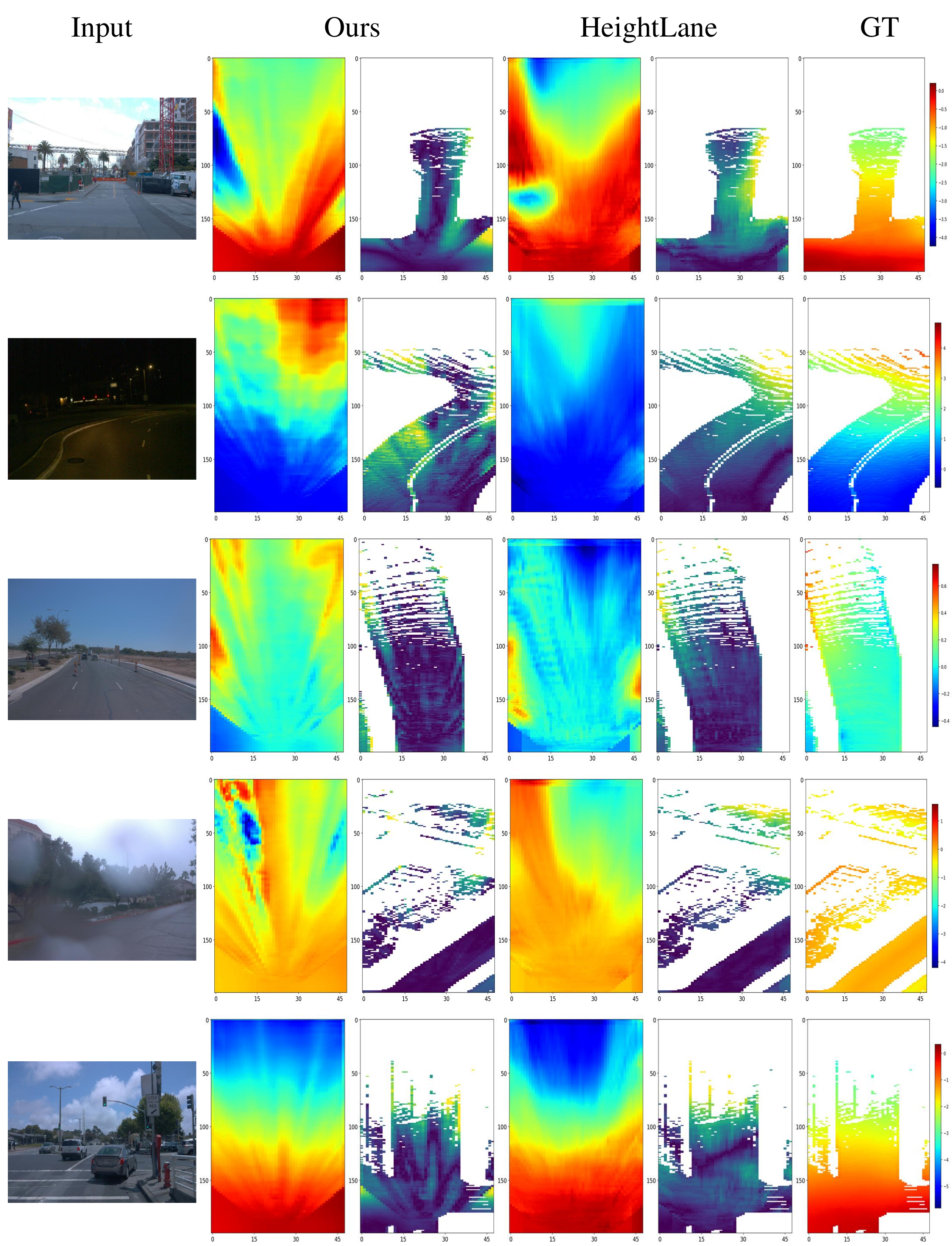} 
  \caption{\textbf{Qualitative comparison of height map estimation on OpenLane Benchmark~\cite{chen2022persformer}.} From left to right: input image $\mathcal{I}$, estimated height maps $\mathcal{H}$ and their corresponding error maps (masked by valid GT regions) for our method and HeightLane~\cite{park2025heightlane}, alongside the GT height map.}
  \label{fig:sp_h_2}
\end{figure}

\begin{figure}[t]
  \centering
  \includegraphics[width=1.0\textwidth]{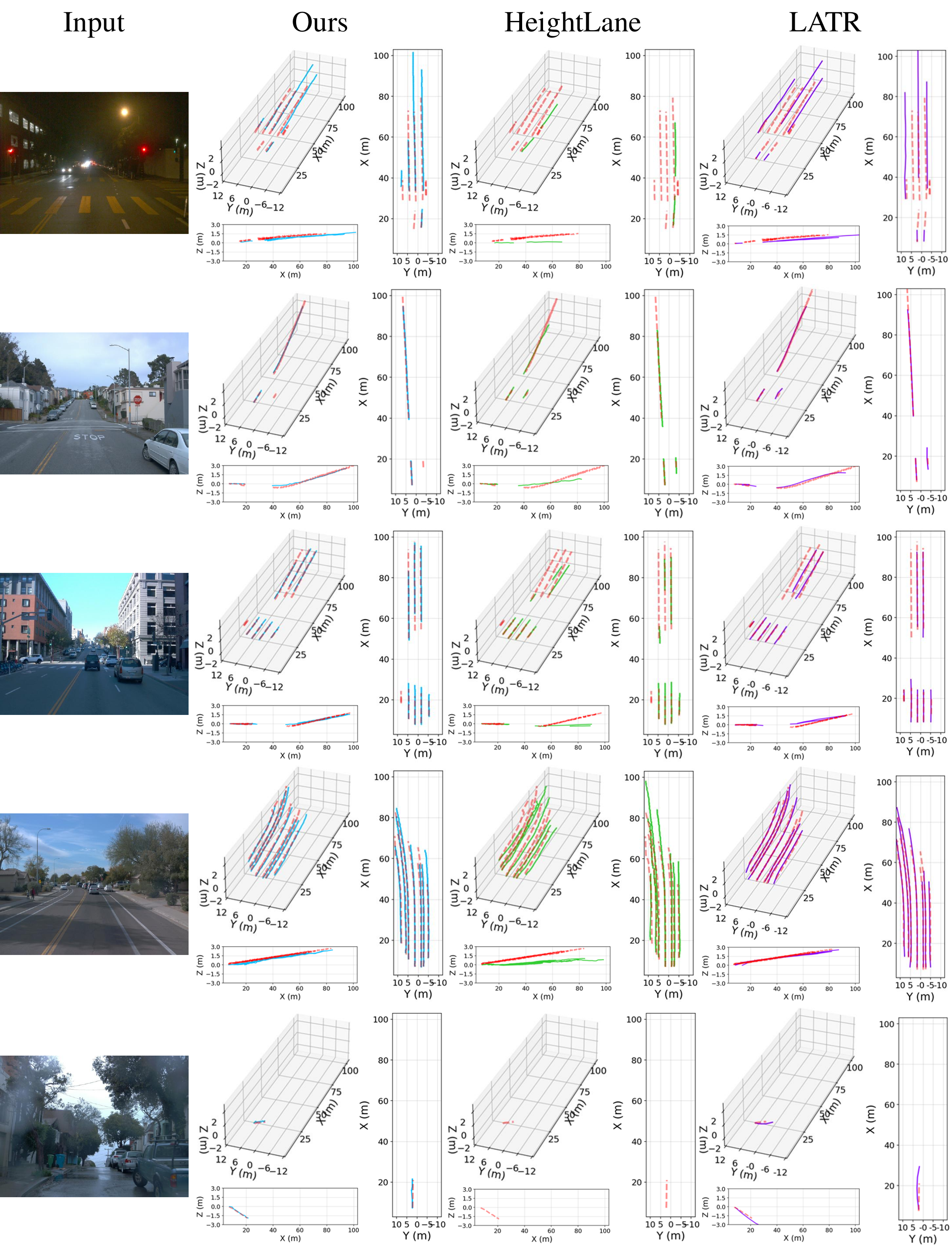} 
  \caption{\textbf{Qualitative result of 3D lane detection on OpenLane Benchmark~\cite{chen2022persformer}.} From left to right: input image $\mathcal{I}$, and 3D lane predictions of \textcolor{blue!60}{HSDF-Lane (ours)}, \textcolor{green}{HeightLane}~\cite{park2025heightlane}, and \textcolor{purple!60}{LATR}~\cite{luo2023latr}. The \textcolor{red}{ground truth (GT)} is indicated by red dashed lines.}
  \label{fig:sp_q_3}
\end{figure}

\begin{figure}[t]
  \centering
  \includegraphics[width=1.0\textwidth]{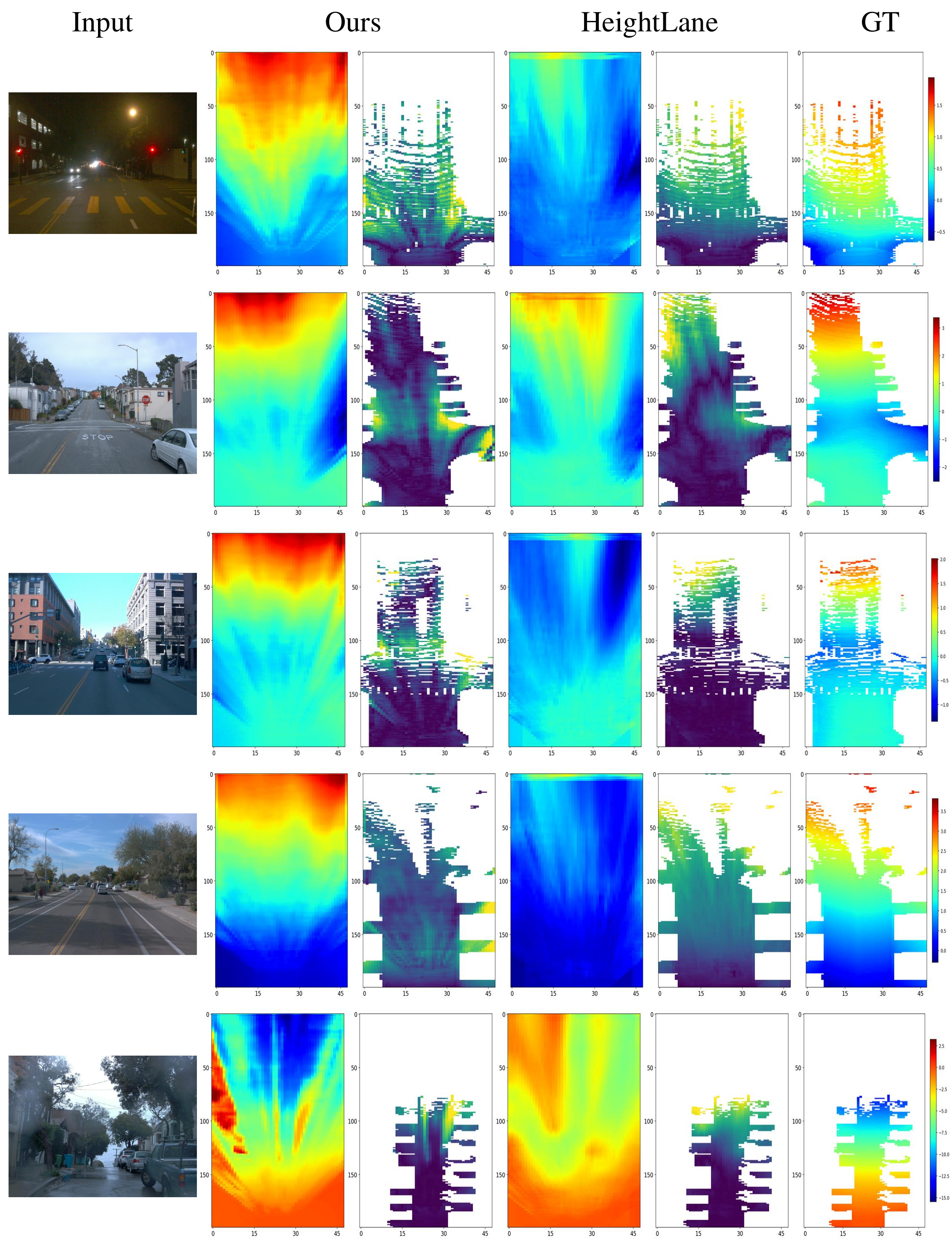} 
  \caption{\textbf{Qualitative comparison of height map estimation on OpenLane Benchmark~\cite{chen2022persformer}.} From left to right: input image $\mathcal{I}$, estimated height maps $\mathcal{H}$ and their corresponding error maps (masked by valid GT regions) for our method and HeightLane~\cite{park2025heightlane}, alongside the GT height map.}
  \label{fig:sp_h_3}
\end{figure}

We provide additional qualitative results on the OpenLane dataset~\cite{chen2022persformer}. Each example consists of a pair of figures showing the 3D lane detection result and the corresponding estimated height map: (\Fref{fig:sp_q_1}, \Fref{fig:sp_h_1}), (\Fref{fig:sp_q_2}, \Fref{fig:sp_h_2}), and (\Fref{fig:sp_q_3}, \Fref{fig:sp_h_3}).
As shown in the figures, our proposed method demonstrates robust performance in both 3D lane detection and height estimation across diverse and challenging driving scenarios. Specifically, the model effectively handles conditions with limited visibility, such as nighttime and adverse weather. Furthermore, it accurately captures complex road geometries, maintaining high precision on severe slopes and sharp curves.

\end{document}